\pdfoutput=1
\documentclass[10pt,twocolumn,letterpaper]{article}

\usepackage[accsupp]{axessibility}
\usepackage{cvpr}              %

\pdfoutput=1
\usepackage[dvipsnames, table]{xcolor}
\usepackage{color}
\usepackage{tikz,pgfplots,pgfplotstable}
\usepackage{float}
\usepackage{amsthm}
\usepackage{diagbox}
\usepackage{xspace}
\usepackage{multirow}
\usepackage{tabularx}
\usepackage{xr}
\usepackage{algorithm} 
\usepackage{algorithmic}  
\usepackage[algo2e]{algorithm2e} 
\usepgfplotslibrary{groupplots}

\definecolor{ClrHighlight}{RGB}{225, 225, 225}
\definecolor{green01270}{RGB}{80, 200, 120}
\definecolor{ForestGreen}{RGB}{34,139,34}
\definecolor{darkgray}{RGB}{113, 121, 126}

\newcommand{\better}[1]{\textcolor{ForestGreen}{#1}}
\newcommand{\worst}[1]{\textcolor{red}{#1}}

\definecolor{cvprblue}{rgb}{0.21,0.49,0.74}
\definecolor{verylightgray}{RGB}{230, 230, 230}

\usepackage[pagebackref,breaklinks,colorlinks,citecolor=cvprblue]{hyperref}

\usepackage{pifont}

\def\paperID{5} %
\def\confName{CVPR}
\def\confYear{2024}

\title{Improving the Robustness of 3D Human Pose Estimation: \\ A Benchmark Dataset and Learning from Noisy Input}

\author{Trung-Hieu Hoang$^1$ \qquad Mona Zehni $^1$ \qquad  Huy Phan$^2$ \qquad Duc Minh Vo$^3$ \qquad Minh N. Do$^1$\\
$^1$University of Illinois at Urbana-Champaign, USA\\
$^2$VinUniversity, Ha Noi, Vietnam \qquad $^3$The University of Tokyo, Japan\\
{\tt\small \{hthieu,mzehni2,minhdo\}@illinois.edu, 20huy.pn@vinuni.edu.vn, vmduc@nlab.ci.i.u-tokyo.ac.jp
}
}

\newrobustcmd*{\mycircle}[1]{\tikz{\filldraw[draw=#1,fill=#1] (0,0) circle [radius=0.075cm];}}
\newrobustcmd*{\mytriangle}[1]{\tikz{\filldraw[draw=#1,fill=#1] (0,0) --
(0.2cm,0) -- (0.1cm,0.2cm);}}

\begin{document}
\maketitle

\begin{abstract}
Despite the promising performance of current 3D human pose estimation techniques, understanding and enhancing their robustness on challenging in-the-wild videos remain an open problem. In this work, we focus on building robust 2D-to-3D pose lifters. To this end, we develop two benchmark datasets, namely Human3.6M-C and HumanEva-I-C, to examine the resilience of video-based 3D pose lifters to a wide range of common video corruptions including temporary occlusion, motion blur, and pixel-level noise. 
We demonstrate the poor generalization of state-of-the-art 3D pose lifters in the presence of corruption and establish two techniques to tackle this issue. First, we introduce Temporal Additive Gaussian Noise (TAGN) as a simple yet effective 2D input pose data augmentation. Additionally, to incorporate the confidence scores output by the 2D pose detectors, we design a confidence-aware convolution (CA-Conv) block. Extensively tested on corrupted videos, the proposed strategies consistently boost the robustness of 3D pose lifters and serve as new baselines for future research. 

\end{abstract}
\vspace*{-1\baselineskip}
\section{Introduction}
\vspace*{-0.5\baselineskip}
\label{sec:intro}
Human pose estimation in 3D (3D HPE) from a monocular RGB video is a challenging computer vision task with a wide range of applications in action recognition~\cite{luvizon2018,liu2020_attention}, virtual/augmented reality~\cite{aliakbarian2022_flag}, human-computer interaction~\cite{WANG2021} and healthcare~\cite{hoangzehni2022_dne}, to name a few. 
In this paper, we focus on the robustness of 2D-to-3D HPE that utilize \textit{off-the-shelf 2D keypoint detection} followed by a \textit{2D-to-3D pose lifter} to lift the sequence of detected 2D keypoints to 3D camera space~\cite{pavllo2019_videopose3d, zheng2021_poseformer, zeng2020_srnet}.
While promising results have been shown on standard benchmarks~\cite{h36m_pami, sigal2010_humaneva, mehta2017-3dhp}, with minimal subjects occlusion, the real-world recordings are far from this controlled setting due to the appearance of external objects or improper camera's field of view~\cite{wang2021_robustness, rempe2021humor, sarandiIROSWS18}. We made an important observation that \textit{these models are less robust to occlusions and disruptions in video appearance}. As this is a two-stage process, one might suggest improving the robustness of the 2D pose detectors~\cite{jiang2021skeletor} instead of the 3D lifter. While this is a valid solution, it is not practical in some real-world applications, such as healthcare, where it is computationally expensive or violates patient's privacy when operating on RGB videos~\cite{hoangzehni2022_dne,s21217315}. Hence, this work benefits systems that favor using off-the-shelf 2D pose estimator, operating on a sequence of detected keypoints without touching the RGB videos.

\begin{figure}[t]
    \centering
    \includegraphics[width=0.9\linewidth]{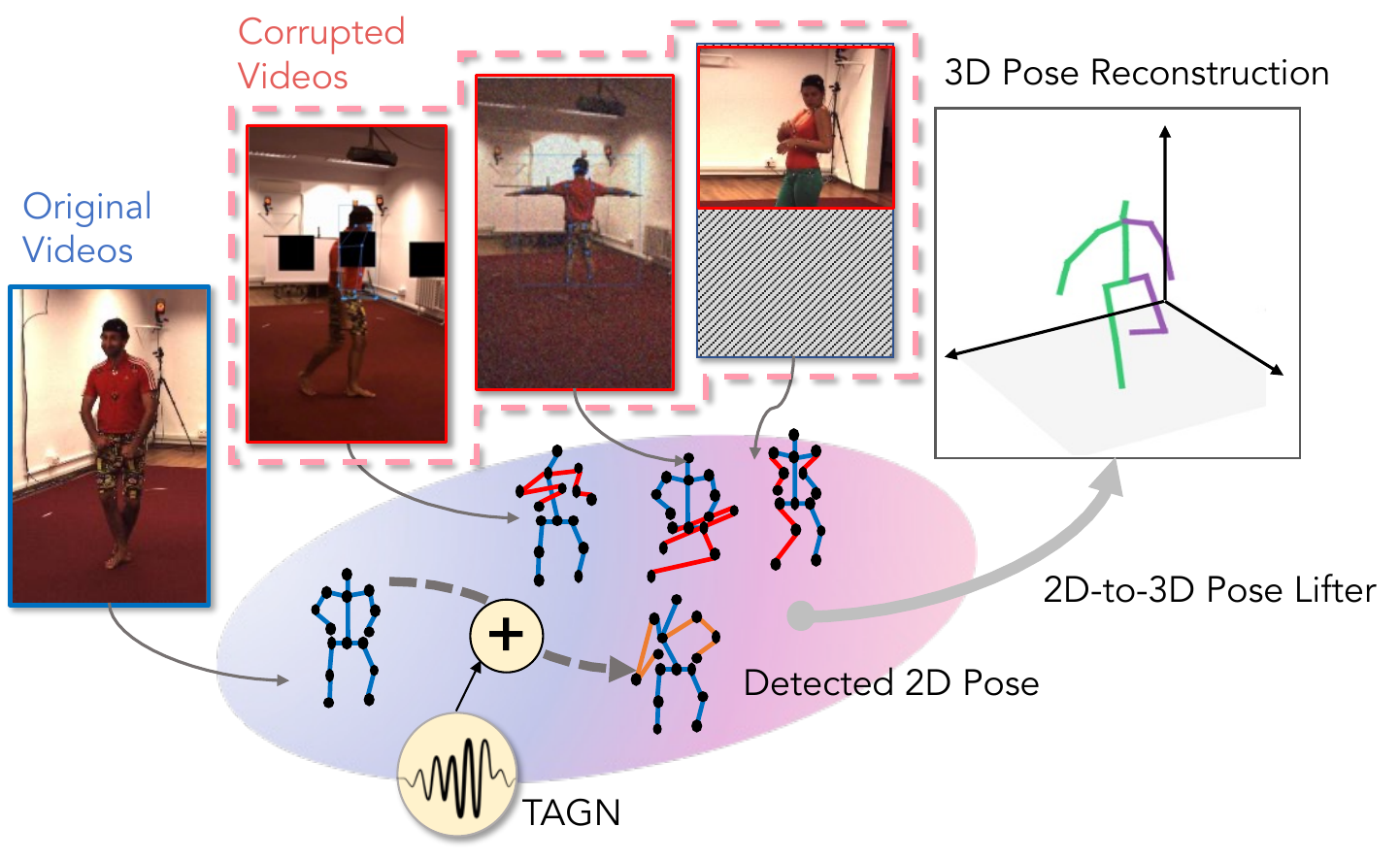}
    \vspace*{-0.5\baselineskip} 
    \caption{Illustration of a 2D-to-3D pose lifter operating on 2D poses detected from corrupted video frames (\textcolor{red}{red} boxes). 
    Our proposed Temporal Additive Gaussian Noise (TAGN) serves as a 2D pose augmentation that adds jitter to the detected 2D poses of the original video frames (\textcolor{blue}{blue} box). TAGN's goal is to improve generalization on test videos with unforeseen visual corruptions.}
    \label{fig:overview}
    \vspace*{-1.5\baselineskip}
\end{figure}

We first systematically study the robustness of 3D HPE by \textit{providing new benchmarks}, namely Human3.6M-C and HumanEva-I-C based on~\cite{h36m_pami, sigal2010_humaneva}.
Following~\cite{wang2021_robustness}, we refer to \textit{corruption} as image-level visual distortions (such as noise, blur, or pixelation). We define frame- and video-level corruption operators inspired by real-world conditions (e.g., partial occlusions or random noises), and use them to construct visually corrupted video datasets. These datasets are obtained after augmenting videos in \cite{h36m_pami, sigal2010_humaneva} with corruptions such as cropping or geometric occlusions. We demonstrate the poor robustness of these models when the test time 2D keypoints are erroneous.

Furthermore, \textit{we introduce two baseline solutions} to enhance the robustness of 3D pose lifters under visual corruption problems in \textit{two scenarios} when the corruptions at test time are known and unknown.
Firstly, we introduce a new \textit{data augmentation} strategy with temporal additive Gaussian noise (TAGN), which can be plug-and-play to any existing methods. TAGN randomly perturbs the 2D keypoints during training, to further simulate 2D input uncertainty and improve the robustness of the pose lifting models at test time. We illustrate the proposed approach in Figure \ref{fig:overview}. 
Secondly, confidence-aware convolution (CA-Conv), an \textit{elaborated version} of the regular temporal convolution block is proposed. While many 2D human pose estimators output a confidence score for each detected keypoint, this information is often discarded when passed to the 3D pose lifter. CA-Conv properly combines the 2D keypoints and their confidence scores to achieve a robust lifting task.
We summarize our main contributions as below:
\begin{itemize}
    \item We propose the first \textit{synthesized video-based 3D HPE datasets} (Human3.6M-C and HumanEva-I-C) explicitly designed to evaluate the robustness of 3D pose lifters. We also introduce a proper MPJPE-based evaluation metric to quantify the robustness and extensively analyze the performance of several state-of-the-art 2D-to-3D pose lifting models~\cite{pavllo2019_videopose3d,zheng20213d,zeng2020_srnet,liu2020_attention,shan2021improving}. 
    \item We develop TAGN, a simple yet\textit{ effective augmentation strategy} that can be conveniently applied while training any 2D-to-3D pose lifter. Unlike image-based augmentations, TAGN is applied to the input 2D keypoints, leading to better test-time generalization than models trained on uncorrupted datasets.
    \item We present CA-Conv, a \textit{confidence-aware extension} of the regular temporal convolution block, and show its efficacy when trained on corrupted videos. To the best of our knowledge, this is the first 3D pose lifter consuming the confidence score associated with the input 2D pose. 
\end{itemize}

\section{Related Works}
\label{sec:related-works}
\noindent\textbf{3D Human Pose Estimation:}  We focus our review on methods devoted to single-person 3D HPE tasks from monocular RGB images and videos. This problem is inherently complex due to depth ambiguity and occlusions. 
We study 3D HPE under two broad categories. One line of work learns to directly estimate the 3D pose in an end-to-end manner from the 2D RGB image with no intermediate supervision~\cite{li2014,park2016,tekin2016direct,pavlakos2017coarse,zhou2016sparseness, sun2018integral,Tekin2017LearningTF,park20163d,Tome_2017_CVPR,sharma2019monocular,mono-3dhp2017,zhou2017towards}. 
In the second class of 3D HPE methods, 2D poses (i.e., detected 2D keypoints of human body joints) are firstly estimated given RGB images~\cite{openpose,wei2016cpm} and subsequently lifted to 3D.
While direct regression-based 3D HPE methods have degraded performance compared to 2D-to-3D lifting due to the lack of intermediate supervision, the main challenge for the lifter models is associated with the inaccuracies of the detected 2D pose~\cite{pavllo2019_videopose3d,martinez2017simple}. Among pioneering works, Chen \textit{et al.}~\cite{Chen_2017_CVPR} approached the 2D-to-3D pose lifting problem as nearest neighbor matching from large 3D mocap datasets. Martinez \textit{et al.}~\cite{martinez2017simple} trained a deep network with linear layers to predict 3D pose from estimated 2D pose. Other 2D-to-3D pose lifting approaches estimate distance matrices~\cite{moreno2017}, employ evolutionary and differentiable data augmentations~\cite{Li_2020_CVPR,gong2021poseaug}, or devise self-supervised methods~\cite{wandt2022elepose}. For 3D pose lifting in videos, the sequential nature and the temporal dependence of the frames are taken into account by processing a temporal sequence of the 2D pose. Temporal dilated convolutions~\cite{pavllo2019_videopose3d}, graph convolution networks~\cite{wang2020motion,zeng2021learning,hu2021conditional} and transformer-based models with temporal and spatial attentions~\cite{zheng20213d,li2022exploiting,jiang2021skeletor} are among the current video-based 3D pose lifting solutions.  

\noindent\textbf{Robust 3D Human Pose Estimation:} Robustness to noise and occlusion is a desired property for any real-world ready 3D HPE model. Recent occlusion-aware or noise-resilient 3D HPE solutions are either semantically occlusion-aware depth-based methods~\cite{7301338} or well-designed convolutional models with occlusion-aware heatmaps~\cite{kocabas2021pare, 9157481}. However, these methods operate on single images or depth maps and are not designed for a video setting. 3D human pose modeling with partial visibility especially for consumer videos~\cite{rockwell2020full, biggs20203d} and depth-aware techniques for multi-person pose estimation~\cite{sun2021monocular, sun2022putting}, are among video-based HPE solutions tackling the robustness aspect. Also,~\cite{9010921} addressed scenarios with self-occlusion by introducing a cylinder man model for pose regularization and occlusion augmentation while adopting optical flow on 2D keypoint heatmaps. \cite{cheng20203d} considered various frame- or keypoint-level occlusion augmentations on 2D heatmaps. Furthermore, \cite{rempe2021humor} builds a motion prior and solves a test-time optimization to find plausible 3D poses coherent with the 2D non-occluded keypoints. Although effective, these methods can be computationally expensive and unsuitable for real-time applications. In addition,~\cite{shan2021improving} incorporated self-supervised pre-trained denoising auto-encoders in a 2D-to-3D human pose estimator. However, the improvement in robustness on corrupted 2D keypoints has not been explicitly shown.

\noindent \textbf{Learning with Additive Jitter}: Training with jittered input is a simple regularization strategy in machine learning that has been shown to improve generalization \cite{reed1992_jittered}. By evading overfitting, it also enhances the robustness and accuracy of neural networks \cite{czarnecki2013_uncertaintymeasure}. Besides, in the field of interpretable machine learning, it has proven helpful in raising the awareness of models' explanations of the input uncertainty \cite{wang2021_show}.

\noindent \textbf{Corrupted Dataset for 3D HPE}: Corrupted datasets are widely used to analyze the robustness of machine learning models in many tasks such as image classification~\cite{hendrycks2018benchmarking}, object detection~\cite{michaelis2019dragon}, semantic segmentation \cite{kamann2021_semantic}, or 2D HPE~\cite{wang2021_robustness}. Meanwhile, occlusion is an HPE domain-specific challenge. S\'{a}r\'{a}ndi \textit{et al.} in \cite{sarandiIROSWS18} introduced geometric occlusions in various shapes such as circles, rectangles, or bars on Human3.6M dataset~\cite{h36m_pami} and showed these augmentations improve test-time performance. More realistic copy-paste augmentations to synthesize occlusion were developed in~\cite{9506392, sarandi2018synthetic}. However, to the best of our knowledge, no existing work studies the robustness of lifter-based 3D HPE models on synthetically corrupted videos. 

\noindent\textbf{Our Work}:  Compared to prior works with single-frame synthesized occlusions \cite{sarandi2018synthetic, sarandiIROSWS18,wang2021_robustness}, 
our benchmark is more comprehensive by including both complex and realistic disruptions like temporal occlusion or motion blur that better match the temporal nature of video data. The closest to our corrupted dataset is \cite{wang2021_robustness}, devoted to the single image 2D HPE task. 
Meanwhile, we focus on the robustness of \textit{temporal} 3D pose lifters~\cite{pavllo2019_videopose3d,zheng20213d} with noisy 2D input keypoints from a corrupted video. 
This enables us to better study temporal occlusions and their impact. Furthermore, we provide solutions to enhance robustness to errors in 2D  pose input, by learning with additive jitter, and taking into account the confidence score of the detected 2D poses when lifting.

\vspace*{-0.2\baselineskip}
\section{Notations and Problem Setup}
\vspace*{-0.3\baselineskip}
\label{sec:background}
\begin{figure}[t]
    \centering
    \includegraphics[width=.95\linewidth]{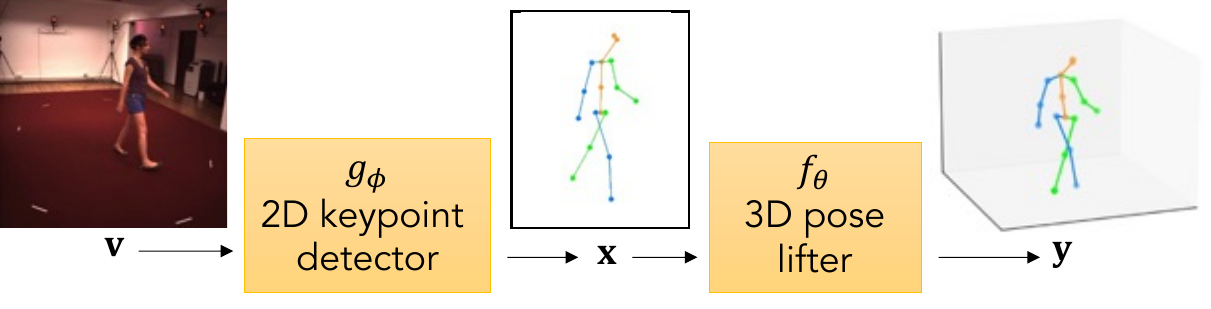}
    \vspace*{-0.5\baselineskip}
    \caption{3D human pose estimation in a 2D-to-3D pose lifting pipeline. The detected 2D pose by $g_\phi$ is lifted to 3D by $f_\theta$.}
    \label{fig:cascade_network}
    \vspace*{-\baselineskip}
\end{figure}

\textbf{Notations:} Consider a two-stage 3D HPE solution as illustrated in Figure~\ref{fig:cascade_network}. Let $g_\phi$ and $f_\theta$ (parameterized by $\phi$ and $\theta$) denote the 2D keypoint detection (such as OpenPose~\cite{openpose} or HRNet~\cite{sun2019_hrnet}) and 3D pose lifter modules, respectively. 
To take into account the temporal nature of the video data, we consider lifter models that consume a sequence of $T$ 2D poses.
A sequence of RGB frames (of $H\times W$ resolution) $\mathbf{v} \in \mathbb{R}^{T \times 3 \times H \! \times \! W}$ is passed as input to $g_\phi$. Next, the 2D detected pose with $J$ keypoints, i.e. $\mathbf{x} = g_\phi(\mathbf{v})$, $\mathbf{x} \in \mathbb{R}^{T \times J \times 2}$, is processed by $f_\theta$ to finally output the estimated 3D pose for the middle frame of the input sequence $\mathbf{y} = f_\theta(\mathbf{x)}$, $\mathbf{y} \in \mathbb{R}^{J \times 3}$. Padding is appropriately applied either at the start or end of the sequence. We \textit{fix} $g_\phi$ to be an off-the-shelf 2D HPE model that is already trained.

\noindent \textbf{Robustness of 3D pose lifters: } We interest in the effect on $f_\theta$ where the \textit{input RGB video frames are randomly corrupted} by a set of image-domain operators (the specific set of operators will be introduced in Sec.~\ref{sec:vd_corruption}), simulating undesirable circumstances that a system may encounter when operating on in-the-wide videos. 
Given a 3D pose lifter trained on the 2D keypoints estimated from uncorrupted frames, if the RGB input is perturbed at test time, the 2D estimated pose by $g_\phi$ would be erroneous. This error would \textit{propagate to the lifter module} $f_\theta$ and leads to poor generalization as expected. 
In some applications where privacy-preserving is crucial, the collected input could be just sequences of 2D detected keypoints from some $g_\phi$ (i.e., improving the robustness of $g_\phi$ is impossible).
Our goal is to train a 2D-to-3D pose lifter, i.e. $f_\theta$, given a sequence of 2D detected keypoints ($\textbf{x}$) by $g_\phi$, that is robust to possible sources of error in the 2D pose input ($\textbf{v}$), such as occlusion or noise.
In previous works~\cite{pavllo2019_videopose3d,zheng20213d}, $f_\theta$ is vulnerable to noisy $\textbf{x}$ since it is trained with ground truth 2D-to-3D pairs.

\noindent \textbf{Scenarios and Proposed Approaches: }
In this work, we propose approaches for two different scenarios: 
\begin{itemize}
    \item \textbf{Scenario 1: } Corruption operators are \textit{unknown}. $f_\theta$ is exposed to an uncorrupted dataset at training while the dataset at testing time is corrupted (i.e., covariate-shift~\cite{10.5555/1462129}). Our data augmentation strategy for increasing the robustness of $f_\theta$ is introduced in Sec.~\ref{sec:tagn}.
    \item  \textbf{Scenario 2: } Corruption operators are \textit{known} during training. The task is to design $f_\theta$ that can efficiently learn from noisy data. We propose a strategy in Sec.~\ref{sec:ca-conv} to replace parts of  $f_\theta$ with our confidence-aware module. 
\end{itemize}

\vspace*{-0.2\baselineskip}
\section{TAGN: Temporal Additive Gaussian Noise}
\begin{figure}[t]
    \centering
    \resizebox{\linewidth}{!}{\input{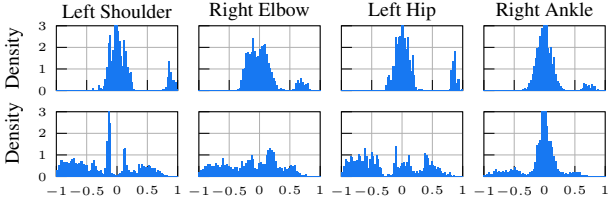}}
    \vspace*{-1.5\baselineskip}
    \caption{Histogram of the $\ell_2$ error of several 2D keypoints detected by HRNet~\cite{sun2019_hrnet}, after applying guided patch erasing \textit{(top)} and Gaussian noise \textit{(bottom)} video corruptions defined in Sec.~\ref{sec:vd_corruption}.}
    \label{fig:noise_histogram}
    \vspace*{-1\baselineskip}
\end{figure}

\vspace*{-0.3\baselineskip}
\label{sec:tagn}
\noindent \textbf{Motivation:}
Under the first scenario, image-domain augmentation is the most direct approach. Since $g_\phi$ is fixed, one can randomly transform the RGB inputs $\mathbf{v}$ by a set of data augmentation operators, mimicking the true corruptions at test time. We then use the 2D detected keypoints of these samples to train $f_\theta$. However, operating on RGB frames is computationally expensive (passing all augmented frames through $g_\phi$), and requires having prior knowledge of the corruptions at test time (to design a proper data-augmentation strategy). Meanwhile, a preferred way is augmenting the lower dimension 2D-pose-domain data ($\textbf{x}$) directly. The drawback is that \textit{it is extremely challenging to model the induced noise in the estimated 2D pose caused by the image-domain corruptions}. This is due to the inherent dependence on the frame appearance, the 2D pose detector, and even the type of video corruption. To support this argument, in Figure~\ref{fig:noise_histogram}, we depict the distribution of the errors in the detected 2D keypoints (by HRNet~\cite{sun2019_hrnet}), after corrupting the videos with two different operators (specifically, guided-patch erasing - \textit{top} and Gaussian noise - \textit{bottom}; see Sec.~\ref{sec:vd_corruption} for further details).  Note how this error distribution varies with the joint and the type of video corruption.

Therefore, we aim to search for an easy-to-simulate and universal noise model, transforming the uncorrupted 2D pose directly. We conjecture that training $f_\theta$ with a simple noisy 2D pose input (without needing to approximate the true induced noise) enhances its robustness and indirectly models detected 2D pose of randomly corrupted frames.

\noindent \textbf{TAGN:} We design TAGN keeping in mind that in real-world settings, for some periods of time, parts of the human body can be occluded, and therefore their corresponding estimated 2D pose is noisy. For example, consider a subject walking around an area with several objects on the ground. While passing nearby objects, the subject's feet may become occluded for a short or extended period. As a result, the error in the estimated 2D keypoints by $g_\phi$ only appears in a portion of 2D joints at some frames. We construct TAGN to closely simulate these situations.
Given a 2D pose sequence, TAGN operates by first randomly selecting $k\%$ of the frames. Next, for each candidate frame, $p\%$ of the joints are drawn randomly and contaminated by additive Gaussian noise with $\mathcal{N}(0, \sigma^2)$ distribution. 
In short, TAGN is a random data augmentation strategy that rather than the RGB input, directly affects the 2D pose input to the lifter. We provide the outline of TAGN in the Supplementary.

Ultimately, one can imagine TAGN to be a simplified error model on the 2D detected keypoint inputs. Despite this, we still found TAGN to be an effective 2D pose data augmentation strategy that improves the robustness to unforeseen video corruptions (as will be shown in Sec.~\ref{ssec:result_tagn}).

\vspace*{-0.5\baselineskip}
\section{CA-Conv: Confidence-aware Convolution}
\vspace*{-0.2\baselineskip}

\label{sec:ca-conv}
\begin{figure}[t]
    \centering

    \begin{tikzpicture}[
    myrectangle/.style={rectangle, draw, minimum width=50, minimum height=40, ultra thick, rounded corners=10, verylightgray, fill=verylightgray}, rec2/.style={rectangle, draw, minimum width=30, minimum height=25, very thick, rounded corners=2, black,fill=lightgray},
    myrectangle2/.style={rectangle, draw, minimum width=60, minimum height=70, ultra thick, rounded corners=10, verylightgray, fill=verylightgray},
    ]
    \node[myrectangle] (a) at (0,0) {};
    \node[rec2] (a) at (0,0) {$\mathbf{w}_\mathbf{s}$};
    \draw[thick,->] (-1.6,0,0) -- (-0.55,0,0) node[left] {};
    \node at (-1,0.15,0) [left] {\small{$\mathbf{s}_\textrm{in}$}};
    \draw[thick,->] (0.55,0,0) -- (1.6,0,0) node[left] {};
    \node at (1.6,0.15,0) [left] {\small{$\mathbf{s}_\textrm{out}$}};

    \node[myrectangle2] (a) at (4,0) {};
    \node[rec2] (a) at (4.3,0.5) {$\mathbf{w}_\mathbf{s}$};
    \node at (3.6,0.5,0) [left] {\large{$\odot$}};
    \draw[thick,-] (2.2,0.5,0) -- (3.18,0.5,0) node[left] {};
    \draw[thick,->] (3.45,0.5,0) -- (3.75,0.5,0) node[left] {};
    \node at (2.8,0.65,0) [left] {\small{$\mathbf{s}_\textrm{in}$}};
    
    \draw[thick,->] (2.2,-0.5,0) -- (3.75,-0.5,0) node[left] {};
    \node[rec2] (a) at (4.3,-0.5) {$\mathbf{w}_\mathbf{c}$};
    \node at (2.8,-0.35,0) [left] {\small{$\mathbf{c}_\textrm{in}$}};
    \draw[thick,->] (3.32,-0.5,0) -- (3.32,0.36,0) node[left] {};
    
    \draw[thick,->] (4.85,0.5,0) -- (5.9,0.5,0) node[left] {};
    \node at (5.85,0.65,0) [left] {\small{$\mathbf{s}_\textrm{out}$}};
    \draw[thick,->] (4.85,-0.5,0) -- (5.9,-0.5,0) node[left] {};
    \node at (5.85,-0.35,0) [left] {\small{$\mathbf{c}_\textrm{out}$}};
    
    \end{tikzpicture}
    \caption{Regular convolution block (left) versus our proposed confidence-aware convolution block (right).}
    \label{fig:ca-conv}
    \vspace*{-1.2\baselineskip}
\end{figure}
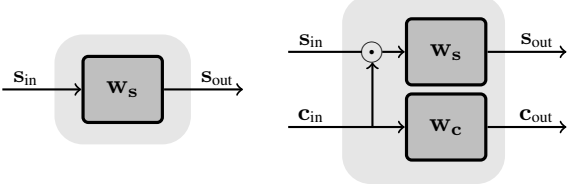

\textbf{Motivation:} Many 2D HPE methods output a confidence score for each detected joint~\cite{openpose,sun2019_hrnet}. However, 3D lifters often \textit{ignore} this score 
and only resort to estimated 2D pose as their input. Here, we aim to incorporate the 2D pose confidence scores in the 3D pose lifting step. To examine this idea, we revise the VideoPose3D~\cite{pavllo2019_videopose3d} architecture by modifying its regular 1D convolution block to be confidence aware, hence the name \textit{confidence-aware convolution} (CA-Conv). The training is performed on noisy keypoints (Scenario 2). This idea can be extended to other architectures.

\noindent \textbf{CA-Conv:} Each CA-Conv block contains two 1D convolutional sub-blocks with separate $\mathbf{w}_\mathbf{s}$ and $\mathbf{w}_\mathbf{c}$ kernels. Furthermore, it consumes two input streams $\mathbf{s}_{\textrm{in}}\in \mathbb{R}^{L_\textrm{in} \times D_{\textrm{in}}}$ and $\mathbf{c}_{\text{in}} \in [0, 1]^{L_{\textrm{in}} \times D_{\textrm{in}}}$, with dimensions $L_{\textrm{in}} \times D_{\textrm{in}}$ (sequence length $\times$ embedding dimension, per sample in each mini-batch). While $\mathbf{s}_{\text{in}}$ is the feature map encoding the pose information, $\mathbf{c}_{\textrm{in}}$ denotes the intermediate features of the confidence scores. The two output streams of each CA-Conv block are obtained following:
\vspace*{-0.5\baselineskip}
\begin{align}
    \mathbf{s}_{\textrm{out}}[d] &\!=\! \text{ReLU}\!\left(\sum_{i=1}^{D_{\textrm{in}}}\!\left(
        \mathbf{s}_{\textrm{in}}[i] \!\odot\! \left(\gamma \!+\! \mathbf{c}_{\textrm{in}}[i] \right)\right) \!*\!  \mathbf{w}_\mathbf{s}[i, d]\right) \label{eq:pose_feature} \\
     \mathbf{c}_{\textrm{out}}[d] &\!=\! \text{Sigmoid}\left(
        \sum_{i=1}^{D_{\textrm{in}}} \mathbf{c}_{\textrm{in}}[i] * \mathbf{w}_\mathbf{c}[i,d]
    \right) \label{eq:conf_feature}
    \vspace*{-0.5\baselineskip}
\end{align}

\noindent where $i$ and $d$ index the input and output channels, respectively. $\mathbf{s}_{\textrm{out}},\mathbf{c}_{\textrm{out}} \in \mathbb{R}^{L_{\textrm{out}} \times D_{\textrm{out}}}$ are output tensors of each CA-Conv block, which are fed to the subsequent blocks. Also, $\odot$ is the element-wise product, while $*$ is the convolution operation. Note that, in~\eqref{eq:conf_feature}, we are channel-wise weighting the pose feature maps $\mathbf{s}_{\textrm{in}}$ by the confidence score features $\mathbf{c}_{\textrm{in}}$. Also, to avoid the effect of 2D pose features with low confidence scores being completely diminished, we consider a positive leakiness term $\gamma$, added to the confidence feature map $\mathbf{c}_\textrm{in}$. Figure~\ref{fig:ca-conv} compares the CA-Conv versus a regular convolution block.

\begin{figure}[t]
    \centering
    \includegraphics[width=.95\linewidth]{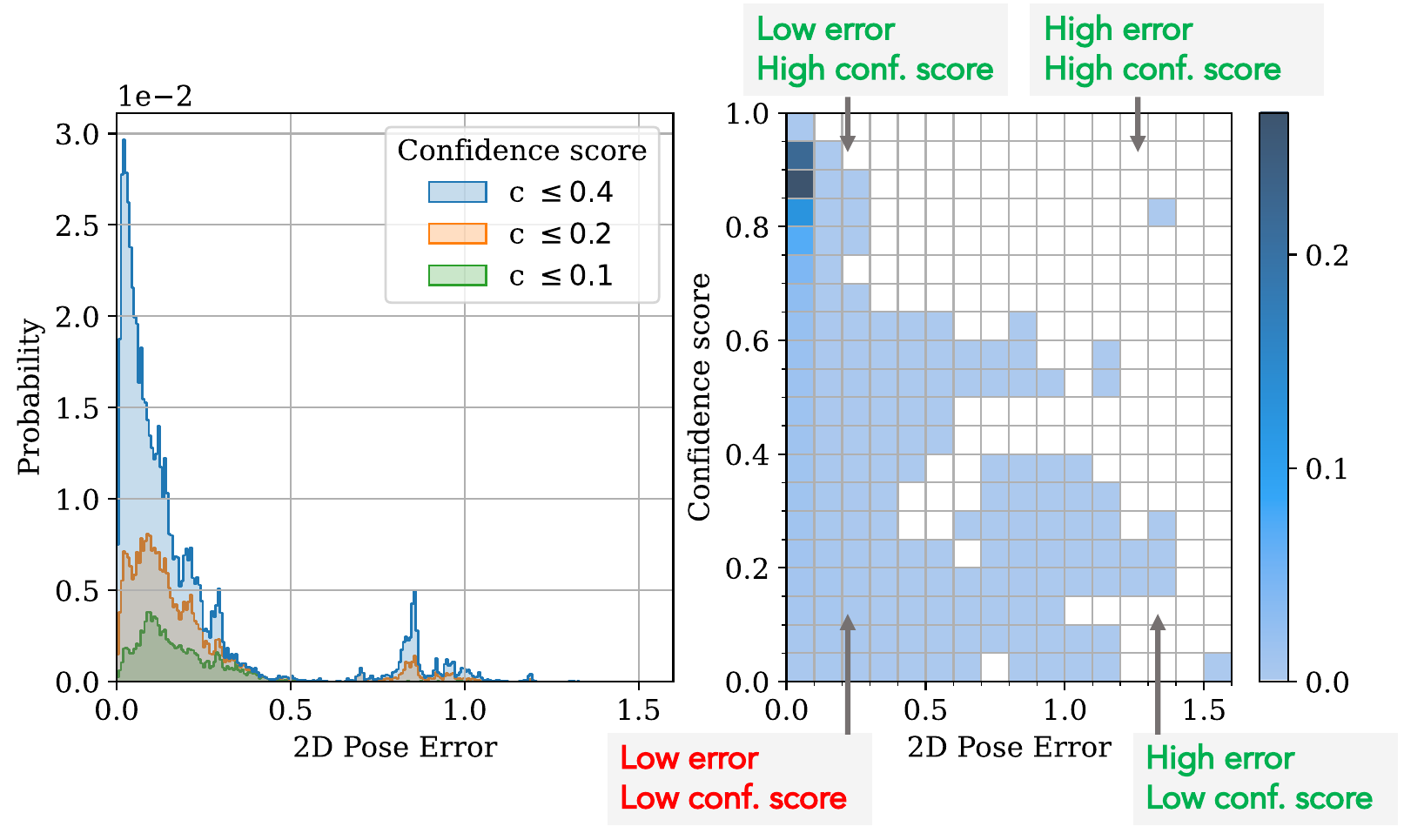}
    \vspace*{-1.0\baselineskip}
    \caption{
    Distribution of the $\ell_2$ error of the left shoulder's detected 2D pose, before and after applying guided patch erasing, for different confidence scores $c$ \textit{(left)}. Joint histogram of the error in the detected 2D pose and the confidence score \textit{(right)}. In the right subplot, the color specifies the density. Best viewed in color.}
    \label{fig:error_histogram}
    \vspace*{-1\baselineskip}
\end{figure}

As there is no explicit supervision on the confidence score when training 2D HPE models, they might be erroneous and thus need to be dealt with cautiously. For instance, we noticed that in HRNet \cite{sun2019_hrnet}, while some 2D keypoints are detected accurately, they might be associated with a relatively low confidence score. To exemplify this, in Figure~\ref{fig:error_histogram}, we visualize the histogram of the $\ell_2$-error between the left-shoulder 2D keypoint output by HRNet, with and without video corruption. We treat the 2D pose with no corruption to be an accurate estimation and compare it with the inferred 2D keypoint in the presence of video corruption.

We observe that the keypoints with high confidence scores ($\geq 0.8$) in Figure \ref{fig:error_histogram}-right, carry useful information with the error distribution concentrated around zero. This means that most of the keypoints that are accurately detected correspond to a high confidence score. On the other hand, the confidence scores might not perfectly correlate with the error in 2D keypoint detection. For example, in Figure \ref{fig:error_histogram}-right, a large portion of 2D keypoint estimations, regardless of their accuracy (i.e. with high and low $\ell_2$ errors) have a low confidence score. Furthermore, in Figure \ref{fig:error_histogram}-left, many less confident keypoint estimations seem to have low detection errors (see the green histogram, concentrated in the low-error region). This implies that ignoring some inferred 2D keypoints simply based on their confidence score can potentially harm the 3D lifter's performance and justifies our leakiness component $\gamma$. 

\label{sec:approaches}

\vspace*{-0.5\baselineskip}
\section{Experimental Setup}
\label{sec:setup}
\vspace*{-0.2\baselineskip}
\subsection{Corrupted 3D HPE Datasets}
\label{sec:vd_corruption}
\vspace*{-0.3\baselineskip}

\noindent \textbf{Standard 3D HPE Benchmarks: } Our evaluation benchmarks are based on the following standard 3D HPE datasets:
\begin{itemize}
    \item \textit{Human 3.6M (H36M)}~\cite{h36m_pami}: This dataset contains 3.6 million video frames capturing 11 subjects performing 15 actions. 
Following \cite{pavllo2019_videopose3d}, our training and test sets include (S1, S5, S6, S7, S8) and (S9, S11) subjects, respectively. 
    \item \textit{HumanEva-I}~\cite{sigal2010_humaneva}: Compared to H36M, this dataset is much smaller, with only 3 subjects captured. We followed the same evaluation protocol as in \cite{pavllo2019_videopose3d, mehta2017_vnect}, where each video is partitioned for training and evaluation purposes. Three actions (Walk, Jog, Box) are used for evaluation.
\end{itemize}

\begin{figure}[t]
    \centering
    \includegraphics[width=\linewidth]{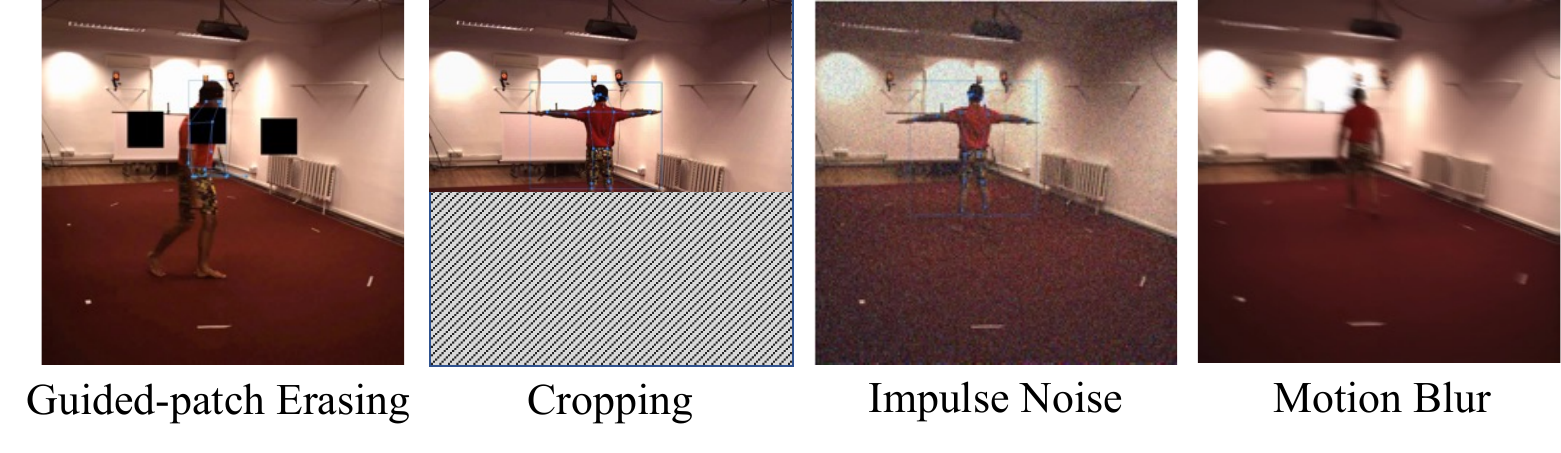}
    \vspace*{-1.5\baselineskip}
    \caption{Examples from our Human3.6M-C dataset. More examples will be provided in the Supplementary.}
    \label{fig:dist_imgs_examples}
    \vspace*{-1.5\baselineskip}
\end{figure}

\noindent \textbf{Image-domain Corruption Operators: }  To propose new benchmarks for evaluating the robustness of video-based 2D-to-3D pose lifters, we introduce several \textit{image-domain operators} that simulate the common artifacts appearing in videos captured in-the-wild.
Following \cite{sarandiIROSWS18, hendrycks2018benchmarking, michaelis2019dragon}, we develop 6 video corruption operators, including guided-patch erasing, temporal patch erasing, pixel-level (Gaussian and impulse) noise, cropping, and synthesized motion blur. Figure~\ref{fig:dist_imgs_examples} illustrates the effect of these operators on example frames from the Human 3.6M~\cite{h36m_pami}. 

In \textit{guided-patch erasing}, instead of regular random-patch occlusion \cite{sarandi2018synthetic} or directly masking the joint positions~\cite{9506392}, we propose a guided erasing strategy to effectively mask out fixed-size square patches in a video. Based on the datasets' ground-truth 2D keypoints, we first uniformly select a subset of joints and track their trajectory throughout a sequence of frames. Next, we choose the masking patch positions such that they have the most overlap with the keypoints' footprint. 
We also consider an extension of this operator called \textit{temporal patch erasing}, where a video is partitioned into $N$ non-overlapping sequences ($N$ is randomly selected from $1$ to $10$) and the guided-patch erasing is applied on each shorter sequence independently. 

To simulate noise introduced due to low-light conditions or video compression~\cite{hendrycks2018benchmarking}, we corrupt the video frames with Gaussian and impulse noises. We also consider \textit{motion blur} operator to account for the possible blurring artifacts due to the motion of the subject or the camera. Finally, to simulate partial visibility of subjects in a video, we adopt a \textit{cropping} operator that horizontally crops the frames. Implementation details are provided in the Supplementary.

\noindent \textbf{Corrupted Variations of H36M \& HumanEva-I}: Applying our corruption operators to standard 3D HPE datasets allows us to generate variations of the original datasets (each with the same number of frames). We call this dataset \textit{H36M-Corrupted} (H36M-C), similar to the naming in~\cite{hendrycks2018benchmarking}. The \textit{test} set of H36M-C is generated by performing {all} video corruptions to the test set of H36M. For the \textit{train} set of H36M-C, each corruption operator is applied to 10\% randomly selected videos from the H36M's training set (without replacement). We keep the remaining unselected videos without corruption in the H36M-C dataset. Therefore, the size of the train sets of H36M and H36M-C datasets is the same.
Similarly, we create the HumanEva-I-C dataset by applying all corruption operators, except the cropping (for reasons we will discuss later), to the whole HumanEva-I.

\subsection{Evaluation Metrics}
\noindent \textbf{MPJPE \& P-MPJPE}: The mean per-joint position error (MPJPE) in mm, computes the Euclidean distance between the ground-truth (GT) and predicted joint positions. P-MPJPE (Procrustes-MJPE), on the other hand, computes MPJPE after the GT and the predicted 3D pose are scale, rotation and translation aligned.

\noindent $\textbf{MPJPE}_{\leq \tau}$: In our corrupted datasets, the corruptions affect each joint differently. To maintain fairness and consistency, MPJPE should only be computed on 3D joints with accurate 2D keypoints as input. For example, if the subject’s legs were masked out, the estimation of the occluded joints is ambiguous and should be excluded from the evaluation. For each frame, we include the $j$-th joint in our evaluation, only if the error on its corresponding 2D keypoints before ($\mathbf{x}$) and after ($\widetilde{\mathbf{x}}$) corruption is less than the threshold $\tau$, i.e. $\Vert \mathbf{x}_j - \widetilde{\mathbf{x}}_j\Vert \leq \tau$. 
We denote this metric as $\text{MPJPE}_{\leq \tau}$. Analogously, we define $\text{P-MPJPE}_{\leq \tau}$. In our experiments, we select $\tau=0.1$, which leads to, on average, $87\%$ of the joints (per frame) being included in the H36M-C test set. We provide the percentage of counted joints in evaluation as a function of $\tau$ in the Supplementary. 
 
\subsection{Baseline Models and Benchmark Performance} 
Our default 2D keypoint detector is HRNet \cite{sun2019_hrnet} trained on COCO \cite{lin2014_coco}. We found HRNet performing better than commonly adopted 2D pose estimators (e.g., CPN~\cite{chen2018_cpn} or Detectron~\cite{wu2019_detectron2}) in 3D pose lifting literature.
For 3D lifters, we experiment with several architectures: VP3D~\cite{pavllo2019_videopose3d}, PoseFormer \cite{zheng2021_poseformer}, SRNet \cite{sun2019_hrnet}, Attention3DHP~\cite{liu2020_attention} and Pose3D-RIE~\cite{shan2021improving}. 
We use the default hyper-parameters and training settings~\cite{sun2019_hrnet,lin2014_coco,wu2019_detectron2,pavllo2019_videopose3d,zheng2021_poseformer,liu2020_attention,shan2021improving}. Models are evaluated on the test sets of H36M-C and HumanEva-I-C. 

For each lifter, we train two models (from scratch) on (1) the original (H36M, HumanEva-I) and (2) the corrupted (H36M-C, HumanEva-I-C) datasets, serving as baselines for the \textit{two scenarios} (Sec.~\ref{sec:background}), respectively.  
In the first scenario, a model is trained on the original dataset with no video corruptions, it generalizes poorly on the corrupted videos. Therefore, we treat its performance as an \textit{upper bound} on MPJPE. On the other hand, the dataset distributions at the training and test time are well-matched for the second scenario. Therefore, its performance is deemed a \textit{lower bound} on MPJPE. For the sake of brevity, we refer to these \textit{benchmark performances} as upper- and lower-bound throughout the rest of this paper. Our approaches are compared against these bounds. A similar extension can be made for lower- and upper-bounds of other metrics such as P-MPJPE. 
To improve the robustness of the models trained on the original dataset, we adopt TAGN as a 2D pose augmentation. We assess these models against the \textit{upper bound}. We also test the effectiveness of CA-Conv for models trained on corrupted videos and compare them with the \textit{lower bound}. 
The implementation details of the methods will be elaborated upon in detail in the Supplementary.

\vspace*{-0.5\baselineskip}
\section{Results}
\label{sec:results}

\subsection{Learning with Jittered 2D Pose}
\label{ssec:result_tagn}

\begin{table*}[t!]
  \caption{$\text{MPJPE}_{\leq 0.1}$ (performance gain compared to the upper bound, lower is better)  evaluated on H36M-C. For all tables in the rest of this
paper, the mean value across $5$ runs is reported for models with TAGN, our methods are highlighted in gray.}
  \vspace*{-0.5\baselineskip}
  \label{tab:mpjpe-tagn}
  \centering
    \resizebox{.95\linewidth}{!}{
    \begin{tabular}{@{}l|cccccc|c@{}}
    \toprule
    \multicolumn{1}{c|}{\textbf{Model}} & \textbf{\begin{tabular}[c]{@{}c@{}}Gaussian \\ Noise\end{tabular}} & \textbf{\begin{tabular}[c]{@{}c@{}}Impulse \\ Noise\end{tabular}} & \textbf{\begin{tabular}[c]{@{}c@{}}Temporal-patch \\ Erasing\end{tabular}} &  \textbf{\begin{tabular}[c]{@{}c@{}}Guided-patch \\ Erasing\end{tabular}} & \textbf{Cropping} & \textbf{\begin{tabular}[c]{@{}c@{}}Motion \\ Blur\end{tabular}} & \textbf{Average} \\ \midrule
    VP3D\cite{pavllo2019_videopose3d}/H36M \textit{(upper bound)}    &  94.27& 96.64 & 99.32 & 116.54 & 118.08 &  78.14 & 100.50\\
    VP3D\cite{pavllo2019_videopose3d}/H36M + {Median Filter}         & 89.55 \better{($\downarrow 4.72$)} & 92.71 \better{($\downarrow 3.93$)} & 97.87 \better{($\downarrow 1.45$)} & 111.86 \better{($\downarrow 4.86$)} & 118.82 \better{($\uparrow 0.78$)}& 75.22 \better{($\downarrow 2.92$)} & 97.65 \better{($\downarrow 2.85$)}\\
    \hline\rowcolor{ClrHighlight}
    VP3D\cite{pavllo2019_videopose3d}/H36M+{TAGN} ($\sigma=0.05; p=k=20\%$) & 91.73 \better{($\downarrow 2.54$)} & 94.12 \better{($\downarrow 2.52$)} & 96.08 \better{($\downarrow 3.24$)} & 112.72 \better{($\downarrow 3.82$)} & 117.23 \better{($\downarrow 0.85$)} & 76.23 \better{($\downarrow 1.91$)} & 98.07 \better{($\downarrow 2.43$)} \\ \rowcolor{ClrHighlight}
    VP3D\cite{pavllo2019_videopose3d}/H36M+{TAGN} ($\sigma=0.1; p=k=30\%$) & 88.44 \better{($\downarrow 5.83$)} & 90.69 \better{($\downarrow 5.95$)} & 93.56 \better{($\downarrow 5.76$)} & 108.95 \better{($\downarrow 7.59$)} & 114.3 \better{($\downarrow 3.78$)} & 75.38 \better{($\downarrow 2.76$)} & 95.22 \better{($\downarrow 5.28$)}\\\rowcolor{ClrHighlight}
    VP3D\cite{pavllo2019_videopose3d}/H36M+{TAGN} ($\sigma=0.3; p=k=50\%$) & 86.3 \better{($\downarrow 7.97$)} & 88.46 \better{($\downarrow 8.18$)} & 92.57 \better{($\downarrow 6.75$)} & 106.74 \better{($\downarrow 9.8$)} & 107.71 \better{($\downarrow 10.37$)} & 75.89 \better{($\downarrow 2.25$)} & 92.94 \better{($\downarrow 7.56$)}\\
    
    VP3D\cite{pavllo2019_videopose3d}/H36M-C \textit{(lower bound)}  &73.11 & 74.03 & 81.65 & 90.56  &  79.76 &  68.40 & 77.92\\
    \hline
    \midrule  %
    PoseFormer\cite{zheng2021_poseformer}/H36M \textit{\textit(upper bound)}                    & 103.16 & 105.09 & 115.91 & 132.37 & 151.62 &  93.13 & 116.88\\\rowcolor{ClrHighlight}
    PoseFormer\cite{zheng2021_poseformer}/H36M+{TAGN} ($\sigma=0.3; p=k=50\%$) & 102.24 \better{($\downarrow 0.92$)} & 104.78 \better{($\downarrow 0.31$)} & 113.8 \better{($\downarrow 2.11$)} & 129.33 \better{($\downarrow 3.04$)} & 148.39 \better{($\downarrow 3.23$)} & 88.7 \better{($\downarrow 4.43$)} & 114.54 \better{($\downarrow 2.34$)} \\
    PoseFormer\cite{zheng2021_poseformer}/H36M-C  \textit{\textit(lower bound)}           & 84.18 & 85.08 & 94.61 & 106.38  &  96.28 &  80.37    & 91.15 \\
    
    \midrule  %
    SRNet\cite{zeng2020_srnet}/H36M \textit{(upper bound)}         & 96.06 & 98.45 & 102.36& 120.29 & 126.08 & 81.27 & 104.08 \\\rowcolor{ClrHighlight}
    SRNet\cite{zeng2020_srnet}/H36M+{TAGN} ($\sigma=0.3; p=k=50\%$))  & 91.68 \better{($\downarrow 4.38$)} & 94.04 \better{($\downarrow 4.41$)} & 96.16 \better{($\downarrow 6.2$)} & 111.94 \better{($\downarrow 8.35$)} & 117.28 \better{($\downarrow 8.8$)} & 78.7 \better{($\downarrow 2.57$)} & 98.3 \better{($\downarrow 5.78$)}\\
    SRNet\cite{zeng2020_srnet}/H36M-C {(lower bound)}      & 78.17 & 79.56 & 84.60 & 93.94  & 82.57  & 71.98 & 81.8 \\
    
    \midrule  %
    Attention3DHP\cite{liu2020_attention}/H36M \textit{(upper bound)}                      & 95.13 & 97.80 & 100.32& 118.48 & 121.02 & 77.48 & 101.7\\\rowcolor{ClrHighlight}
    Attention3DHP\cite{liu2020_attention}/H36M+{TAGN} ($\sigma=0.3; p=k=50\%$) & 92.05 \better{($\downarrow 3.08$)} & 93.94 \better{($\downarrow 3.86$)} & 100.48 \better{($\uparrow 0.16$)} & 115.08 \better{($\downarrow 3.4$)} & 112.8 \better{($\downarrow 8.22$)} & 84.35 \better{($\uparrow 6.87$)} & 99.78 \better{($\downarrow 1.92$)}\\
    Attention3DHP\cite{liu2020_attention}/H36M-C \textit{(lower bound)}    & 73.67 & 74.82 & 91.96 & 91.21  &  79.2 &  68.24 &  78.18 \\
    \midrule
    Pose3D-RIE \cite{shan2021improving}/H36M \textit{(upper bound)} & 104.32 & 105.91 & 108.72 & 120.05 & 117.83 & 94.27 & 108.51\\\rowcolor{ClrHighlight}
    Pose3D-RIE\cite{shan2021improving}/H36M+{TAGN} ($\sigma=0.3; p=k=50\%$) & 92.14 \better{($\downarrow 12.18$)} & 101.34 \better{($\downarrow 4.57$)} & 105.45 \better{($\downarrow 3.27$)} & 115.31 \better{($\downarrow 4.74$)} & 104.42 \better{($\downarrow 13.41$)} & 92.4 \better{($\downarrow 1.87$)} & 101.84 \better{($\downarrow 6.67$)} \\
    Pose3D-RIE \cite{shan2021improving}/H36M-C \textit{(lower bound)} & 86.41 & 87.56 & 93.47 & 97.12 & 94.52 & 83.48 & 90.42\\
    \bottomrule
    \end{tabular}
    }
    \vspace*{-0.5\baselineskip}
\end{table*}

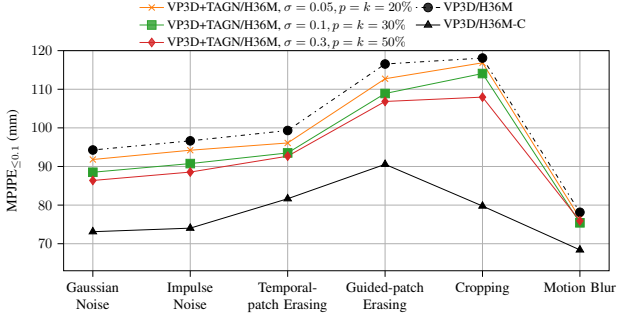
\begin{figure}[t]
    \centering
    \resizebox{\linewidth}{!}{\begin{tikzpicture}

\definecolor{crimson2143940}{RGB}{214,39,40}
\definecolor{darkgray176}{RGB}{176,176,176}
\definecolor{darkorange25512714}{RGB}{255,127,14}
\definecolor{forestgreen4416044}{RGB}{44,160,44}
\definecolor{lightgray204}{RGB}{204,204,204}

\begin{axis}[
legend cell align={left},
legend style={
  fill=none,
  fill opacity=0.8,
  draw opacity=1,
  text opacity=1,
  at={(0.13,1.25)},
  anchor=north west,
  draw=none,
  legend columns=2,
  font=\small
},
height=7cm, width = 15cm,
tick align=outside,
tick pos=left,
x grid style={darkgray176},
xmajorgrids,
xmin=-0.3, xmax=5.3,
xtick style={color=black},
xtick={0,1,2,3,4,5},
xticklabels={Gaussian Noise, Impulse Noise,Temporal-patch Erasing, Guided-patch Erasing,Cropping,Motion Blur},
x tick label style={align = center, text width=2cm},
y grid style={darkgray176},
ylabel={MPJPE$_{\leq 0.1}$ (mm)},
ymajorgrids,
ymin=63.0, ymax=120.0,
ytick style={color=black}
]

\addplot [semithick, darkorange25512714, mark=x, mark size=3, mark options={solid}]
table {%
0 91.81
1 94.21
2 96.09
3 112.72
4 116.85
5 76.62
};
\addlegendentry{VP3D+TAGN/H36M, $\sigma=0.05, p=k=20\%$}
\addplot [semithick, black, dash pattern=on 1pt off 3pt on 3pt off 3pt, mark=*, mark size=3, mark options={solid}]
table {%
0 94.27
1 96.64
2 99.32
3 116.54
4 118.08
5 78.14
};
\addlegendentry{VP3D/H36M }
\addplot [semithick, forestgreen4416044, mark=square*, mark size=3, mark options={solid}]
table {%
0 88.51
1 90.75
2 93.51
3 108.9
4 114.07
5 75.41
};
\addlegendentry{VP3D+TAGN/H36M, $\sigma=0.1, p=k=30\%$}
\addplot [semithick, black, mark=triangle*, mark size=3, mark options={solid,rotate=0}]
table {%
0 73.11
1 74.03
2 81.65
3 90.56
4 79.76
5 68.4
};
\addlegendentry{VP3D/H36M-C}
\addplot [semithick, crimson2143940, mark=diamond*, mark size=3, mark options={solid}]
table {%
0 86.38
1 88.55
2 92.67
3 106.84
4 107.96
5 75.92
};
\addlegendentry{VP3D+TAGN/H36M, $\sigma=0.3, p=k=50\%$}

\end{axis}

\end{tikzpicture}}
    \vspace*{-1.5\baselineskip}
    \caption{Effect of TAGN with different severity levels. We compare against the lower- and upper-bound performances plotted with \mytriangle{black} and \mycircle{black} markers, respectively. TAGN outperforms the upper bound. All models are tested on H36M-C.}
    \label{fig:mpjpe_jittered_data}
    \vspace*{-\baselineskip}
\end{figure}

\noindent \textbf{TAGN with VP3D}: We use TAGN to augment the 2D pose training set of H36M with various ($p$, $k$, $\sigma$) parameters and use this dataset to train a VP3D model (with a temporal receptive field of 27 frames). Note that, TAGN is only applied once at the beginning of the training process. We repeat each TAGN experiment five times with different noise realizations on the 2D pose input and report the average $\text{MPJPE}_{\leq 0.1}$ evaluated on H36M-C in Table \ref{tab:mpjpe-tagn} (rows 1-6). 

We found a small standard deviation on the performance metrics across all runs in most cases (details in the Supplementary). Despite its simplicity, TAGN empirically improves the lifter's robustness across multiple video corruptions. Noteworthy, the lifters trained with noisy 2D poses are universal and lack knowledge of the exact input RGB video corruption operator applied at test time. The gap between the performance of TAGN under different settings versus the lower- and upper-bounds is visualized in Figure~\ref{fig:mpjpe_jittered_data}. We also consider a baseline in which the 2D keypoints of H36M-C are denoised by a median filter with kernel size $5$ (at both train and test times).

Models with TAGN successfully outperform the lower-bound and the simple denoising approach. Besides, the higher level of noise added to the 2D pose during training leads to improved robustness of VP3D models. 
Noteworthy, TAGN does not introduce surplus complexity during training. Unlike video-level corruptions, we can augment any dataset with additive jitter instantaneously, evade heavy computations, or running inference on 2D pose detectors.

\noindent \textbf{TAGN with Other Pose Lifters:} Besides VP3D, in Table~\ref{tab:mpjpe-tagn}, we demonstrate the effect of TAGN on other state-of-the-art 2D-to-3D pose lifting solutions. Given the diversity of the studied architectures, we have the same observation regarding the enhanced robustness after using TAGN.  
\begin{table*}[t!]
  \caption{$\text{MPJPE}_{\leq 0.1}$ and performance gain compared to the upper bound (tested on HumanEva-I-C).}
  \label{tab:mpjpe-humaneva}
  \vspace*{-0.5\baselineskip}
  \centering
    \resizebox{.9\linewidth}{!}{
    \begin{tabular}{@{}l|ccccc|c@{}}
    \toprule
    \multicolumn{1}{c|}{\textbf{Model}}  & \textbf{\begin{tabular}[c]{@{}c@{}}Gaussian \\ Noise\end{tabular}} & \textbf{\begin{tabular}[c]{@{}c@{}}Impulse \\ Noise\end{tabular}} & \textbf{\begin{tabular}[c]{@{}c@{}}Temporal-patch \\ Erasing\end{tabular}} &  \textbf{\begin{tabular}[c]{@{}c@{}}Guided-patch \\ Erasing\end{tabular}} & \textbf{\begin{tabular}[c]{@{}c@{}}Motion \\ Blur\end{tabular}} & \textbf{Average}\\ \midrule
    VP3D\cite{pavllo2019_videopose3d}/HumanEva-I \textit{(upper bound)}               
        & $66.67 $ & $62.05 $ & $33.64 $ & $33.32 $ & $55.82 $ & $50.30 $ \\\rowcolor{ClrHighlight}
    VP3D\cite{pavllo2019_videopose3d}/HumanEva-I+{TAGN}($\sigma=0.3; p=k=10\%$)
        & 59.21 \better{$(\downarrow 7.36)$} & 55.22 \better{$(\downarrow 6.83)$} & 33.74 \worst{$(\uparrow 0.10)$} & 33.39 \worst{$(\uparrow 0.07)$}& 48.87 \better{$(\downarrow 6.95)$} & 46.11 \better{$(\downarrow 4.19)$} 
    \\
    VP3D\cite{pavllo2019_videopose3d}/HumanEva-I-C \textit{(lower bound)} 
        & $35.19 $ & $33.92 $ & $32.33 $ & $31.72 $ & $33.41 $ & $33.31 $  \\
    \bottomrule
    \end{tabular}
    }
\end{table*}

\noindent \textbf{TAGN with VP3D on HumanEva-I}: 
We also evaluate VP3D with TAGN on HumanEva-I \cite{sigal2010_humaneva} in Table \ref{tab:mpjpe-humaneva}. 
Due to the length of videos in this dataset, the distortion ratios (i.e. $p$ and $k$) of TAGN are reduced to $10\%$. Furthermore, we found that cropping severely affects the quality of the estimated 2D pose. We attribute this to the fact that in HumanEva-I, the videos have lower resolution and the subjects remain mainly close to the center of the frame. Therefore, horizontal cropping completely removes visual cues of the lower body joints throughout almost all frames, leading to significant errors in 2D pose estimation. Thus, we exclude this video corruption in our analysis in Table \ref{tab:mpjpe-humaneva}.

TAGN outperforms the upper-bound on three corruptions (motion blur, Gaussian, and impulse noise). However, the performance on temporal- and guided- patch erasing does not improve for the aforementioned reasons (e.g. lower mobility of the subjects). In our analysis on HumanEva-I, we excluded lifters that require a large training set (e.g., PoseFormer, Attention3DHP) or are unadaptable to a different human skeleton layout (e.g., SRNet).

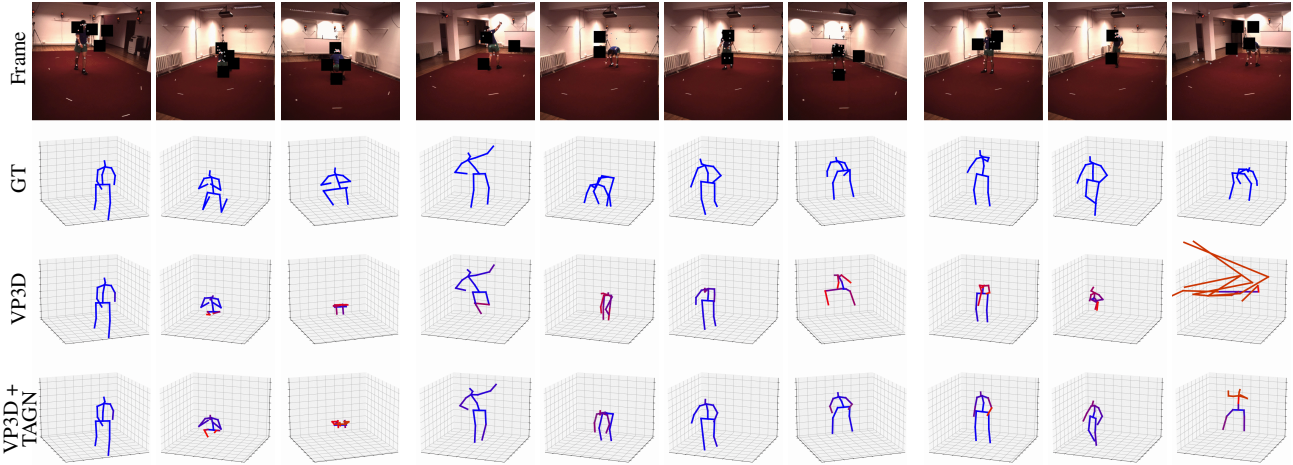
\begin{figure*}[t!]
    \pgfplotsset{every x tick label/.append style={font=\tiny, yshift=0.5ex}}
    \pgfplotsset{every y tick label/.append style={font=\tiny, xshift=0.5ex}}
    \centering
    
        \hspace{-10pt}
        \begin{tikzpicture}
        \draw (0, 0) node[inner sep=0] {\includegraphics[width=1.57cm]{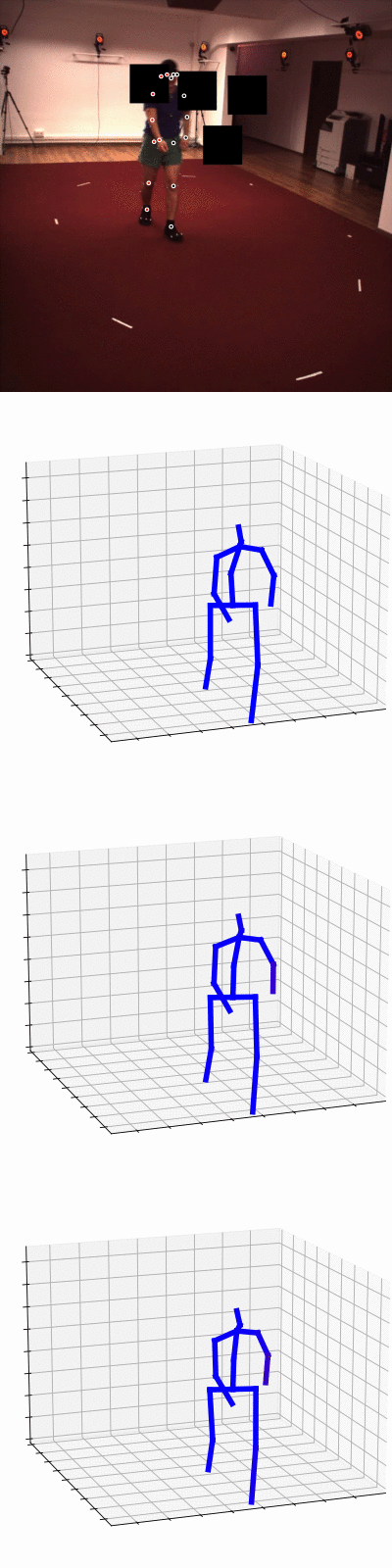}};
        \node at (-1.3,-2.4,0) [below, rotate=90] {\footnotesize{VP3D + }};
        \node at (-1.05,-2.4,0) [below, rotate=90] {\footnotesize{TAGN}};
        \node at (-1.2,-0.8,0) [below, rotate=90] {\footnotesize{VP3D}};
        \node at (-1.2,0.8,0) [below, rotate=90] {\footnotesize{GT}};
        \node at (-1.2,2.4,0) [below, rotate=90] {\footnotesize{Frame}};
        
        \draw (1.65, 0) node[inner sep=0] {\includegraphics[width=1.57cm]{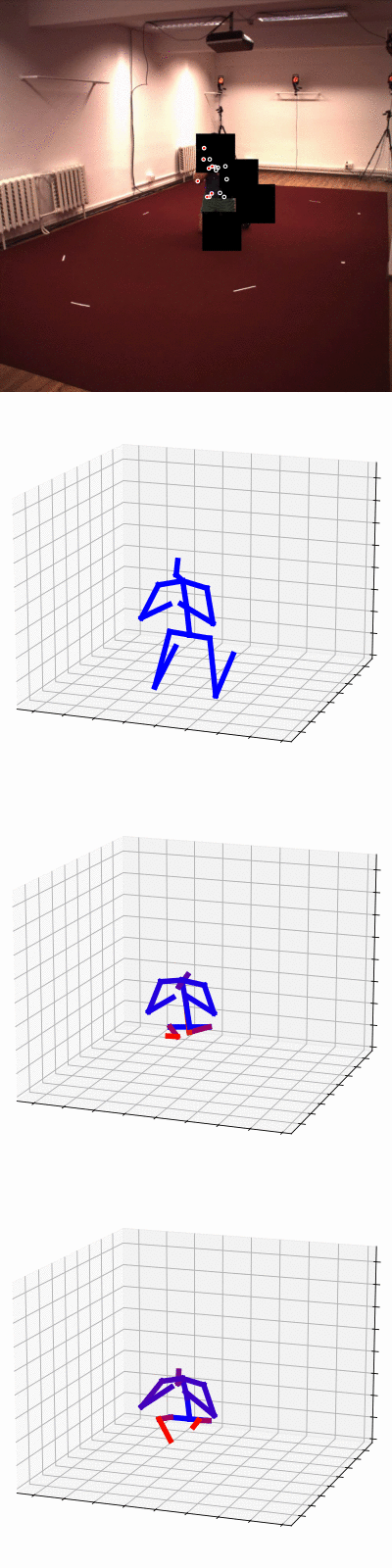}};
        \draw (3.3, 0) node[inner sep=0] {\includegraphics[width=1.57cm]{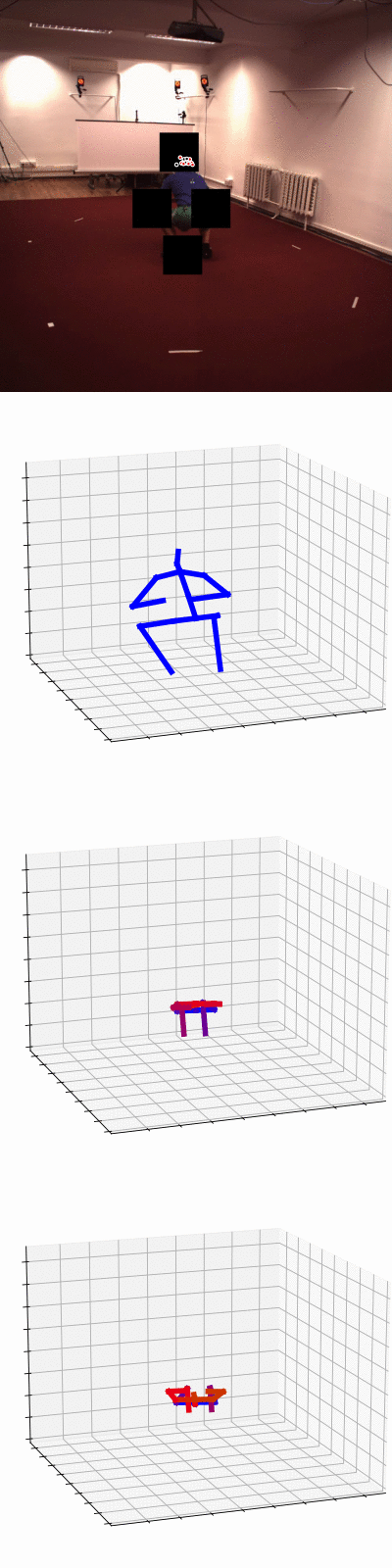}};
        
        \draw (5.1, 0) node[inner sep=0] {\includegraphics[width=1.57cm]{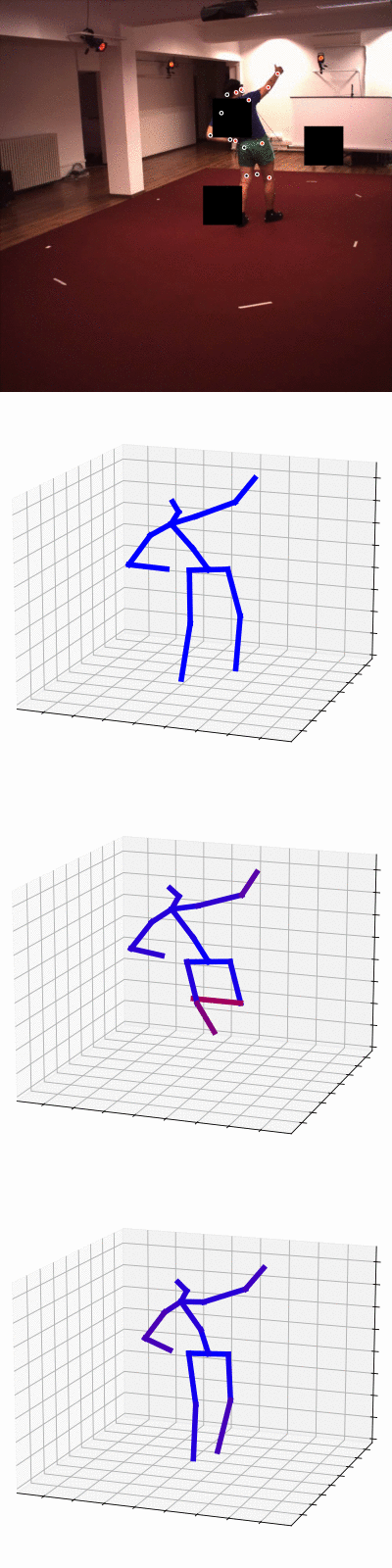}};
        \draw (6.75, 0) node[inner sep=0] {\includegraphics[width=1.57cm]{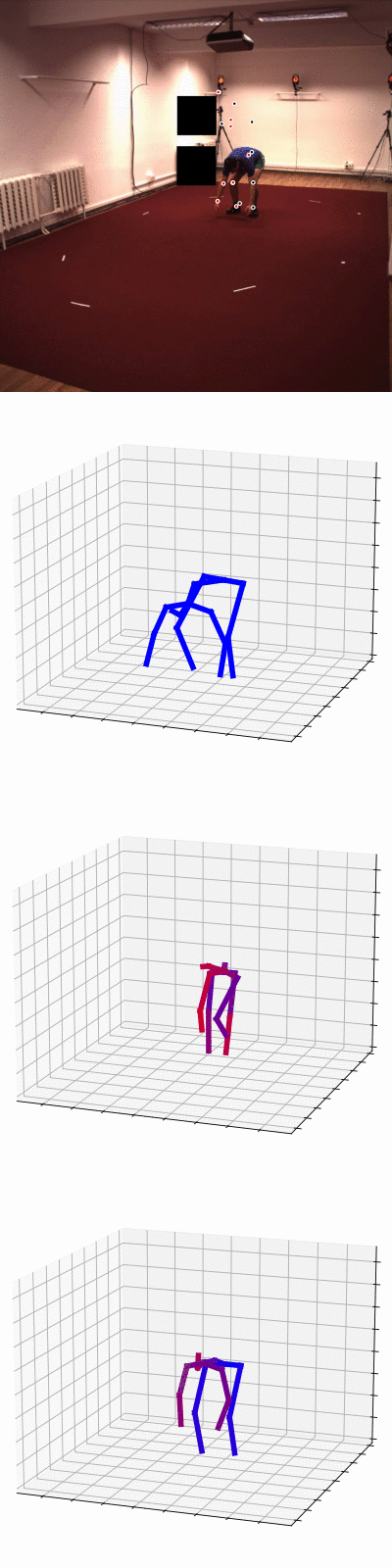}};
        \draw (8.4, 0) node[inner sep=0] {\includegraphics[width=1.57cm]{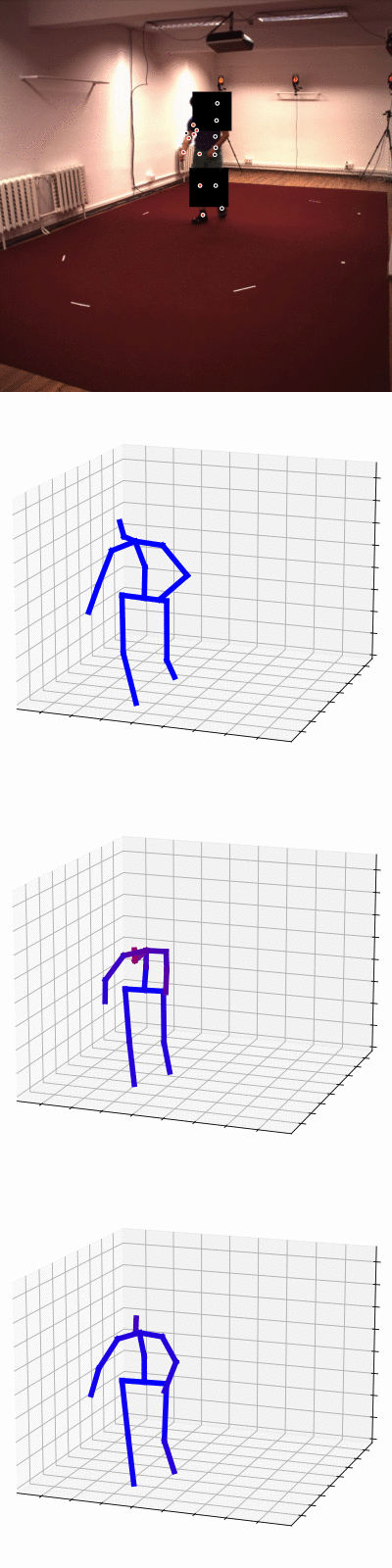}};
        \draw (10.05, 0) node[inner sep=0] {\includegraphics[width=1.57cm]{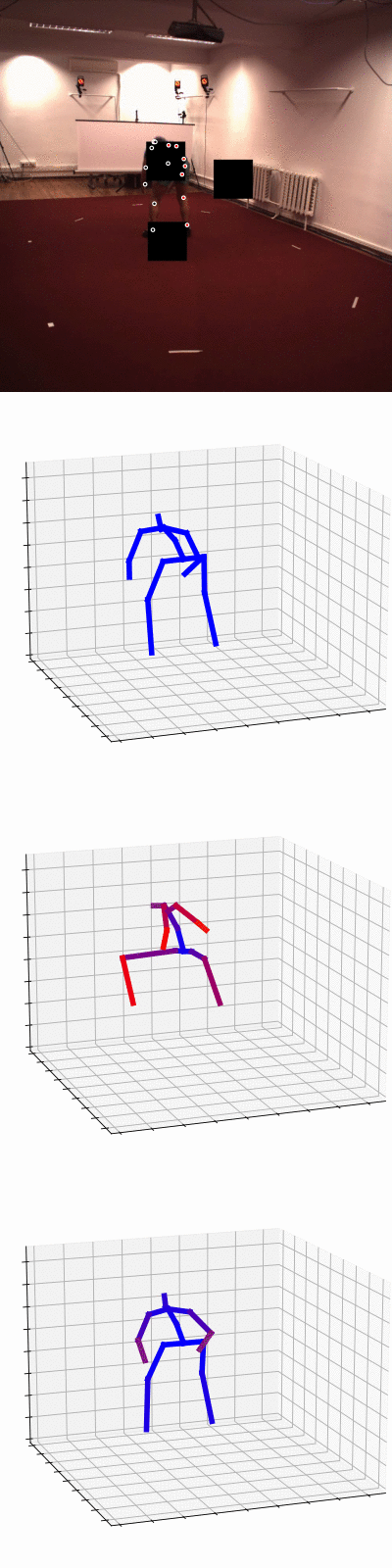}};
        
        \draw (11.85, 0) node[inner sep=0] {\includegraphics[width=1.57cm]{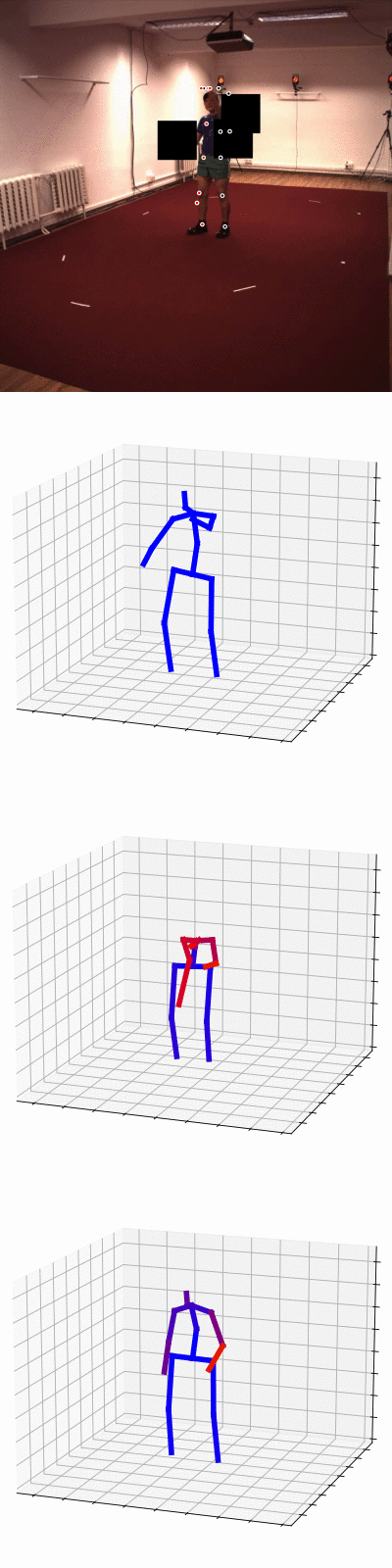}};
        \draw (13.5, 0) node[inner sep=0] {\includegraphics[width=1.57cm]{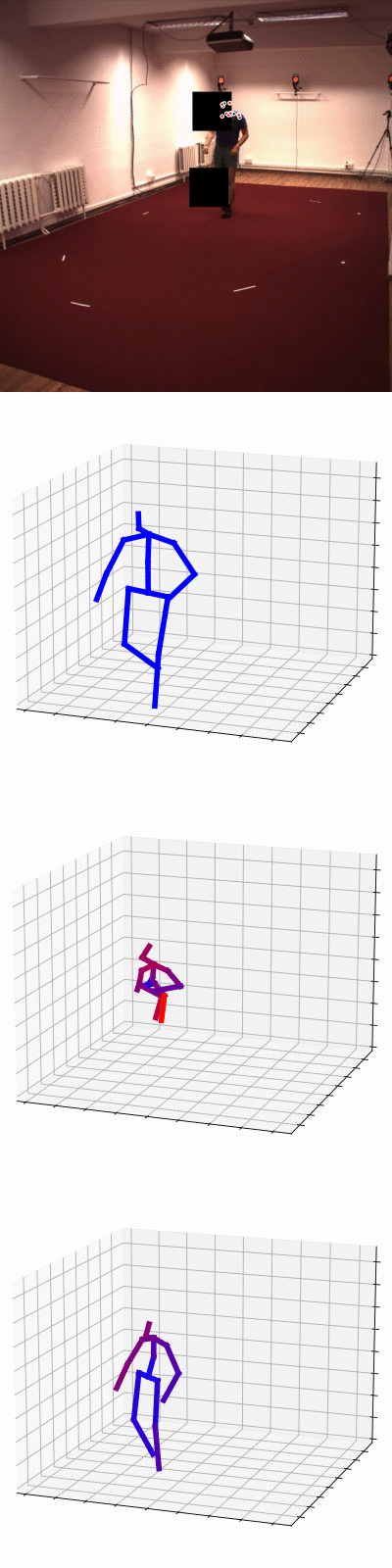}};
        \draw (15.15, 0) node[inner sep=0] {\includegraphics[width=1.57cm]{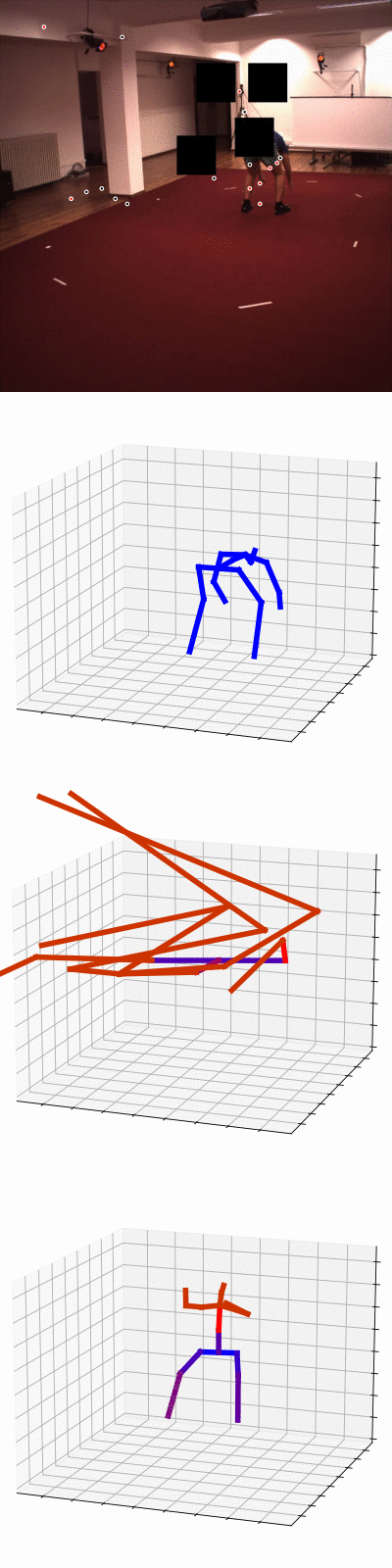}};

        \end{tikzpicture}
    \vspace*{-\baselineskip}
    \caption{Qualitative comparison of VP3D+TAGN ($\sigma=0.3, p=k=50\%$) versus VP3D, on H36M-C dataset (with guided-patch erasing corruption). The bone color leading to a joint turns \textcolor{red}{red} when its MPJPE increases.}
    \vspace*{-1.0\baselineskip}
    \label{fig:tagn-qualitative}
\end{figure*}

\noindent\textbf{Qualitative Comparison:} In Figure \ref{fig:tagn-qualitative}, we visualize the effect of TAGN 2D pose augmentation on the quality of the estimated 3D pose. When the input 2D pose is precise or highly inaccurate (column 1), VP3D and VP3D+TAGN perform similarly (columns 2-3). However, we observe the advantage of TAGN, when the errors in 2D input pose \textit{happen only on a few keypoints or temporarily} (columns 4-7). Remarkably, while VP3D fails in some challenging situations, our VP3D+TAGN still preserves its robustness and outputs a reasonable 3D pose (columns 8-10).

\subsection{Learning with Confidence Scores}
\label{ssec:result_caconv}
\begin{table*}[t!]
  \caption{Effect of CA-Conv blocks on $\text{MPJPE}_{\leq 0.1}$. All models are trained and tested on H36M-C dataset.}
  \vspace*{-0.5\baselineskip}
  \label{tab:mpjpe-ac-conv}
  \centering
    \resizebox{.95\linewidth}{!}{
    \begin{tabular}{@{}l|cccccc|c@{}}
    \toprule
    \multicolumn{1}{c|}{\textbf{Model}} & \textbf{\begin{tabular}[c]{@{}c@{}}Gaussian \\ Noise\end{tabular}} & \textbf{\begin{tabular}[c]{@{}c@{}}Impulse \\ Noise\end{tabular}} & \textbf{\begin{tabular}[c]{@{}c@{}}Temporal-patch \\ Erasing\end{tabular}} &  \textbf{\begin{tabular}[c]{@{}c@{}}Guided-patch \\ Erasing\end{tabular}} & \textbf{Cropping} & \textbf{\begin{tabular}[c]{@{}c@{}}Motion \\ Blur\end{tabular}} & \textbf{Average}\\ \midrule
    VP3D\cite{pavllo2019_videopose3d}/H36M-C \textit{(lower bound)}          & 73.11 & 74.03 & 81.65 & 90.56  &  79.76 &  68.40 & 77.92\\
    VP3D\cite{pavllo2019_videopose3d}/H36M-C + \textit{Median Filter}        & 75.06 \worst{($\uparrow 1.95$)} & 76.14 \worst{($\uparrow 2.11$)} & 82.42 \worst{($\uparrow 0.77$)} & 89.11 \better{($\downarrow 1.45$)} & 82.04 \worst{($\uparrow 2.28$)}& 70.12 \worst{($\uparrow 1.72$)} & 79.14 \worst{($\uparrow 1.22$)}\\
    VP3D\cite{pavllo2019_videopose3d}/H36M-C + \textit{Conf. Concat}         & 74.42 \worst{($\uparrow 1.31$)} & 75.18 \worst{($\uparrow 1.15$)} & 80.81 \better{($\downarrow 0.84$)} & 89.52 \better{($\downarrow 1.04$)} & 81.19 \worst{($\uparrow 1.43$)} & 68.84 \worst{($\uparrow 0.44$)} & 78.33 \worst{($\uparrow 0.41$)}\\
    \midrule\rowcolor{ClrHighlight}
    VP3D\cite{pavllo2019_videopose3d}+{CA-Conv} ($\gamma=0$)/H36M-C  & 73.11 \better{($\downarrow 0.00$)} & 73.96 \better{($\downarrow 0.07$)} & 79.77 \better{($\downarrow 1.88$)} & 88.00 \better{($\downarrow 2.56$)} & 78.83 \better{($\downarrow 0.93$)} & 67.85 \better{($\downarrow 0.55$)} & 76.92 \better{($\downarrow 1.00$)} \\\rowcolor{ClrHighlight}
    VP3D\cite{pavllo2019_videopose3d}+{CA-Conv} ($\gamma=1$)/H36M-C  & 72.31 \better{($\downarrow 0.80$)} & 73.28 \better{($\downarrow 0.75$)} & 79.45 \better{($\downarrow 2.20$)} & 87.72 \better{($\downarrow 2.84$)} & 76.84 \better{($\downarrow 2.92$)} & 67.28 \better{($\downarrow 1.24$)} & 73.64 \better{($\downarrow 1.79$)}\\
    \bottomrule
    \end{tabular}
    }
\vspace*{-0.7\baselineskip}
\end{table*}

In Table~\ref{tab:mpjpe-ac-conv}, we showcase the results of VP3D with all regular 1D convolutions replaced by our CA-Conv blocks (VP3D+CA-Conv). For this analysis, we trained on H36M-C. To the best of our knowledge, there is no similar 2D-to-3D human pose lifting approach that directly consumes the output confidence scores from a 2D pose detector. Thus, we assess the performance of VP3D+CA-Conv against two baselines. Notably, we show that a simple extension of VP3D - concatenating the confidence score to the 2D pose input (VP3D+{Conf.-Concat}) similar to \cite{yan2018_stgc} or denoising the 2D keypoints sequence with median filter (with kernel size $5$) before passing to the lifter are not effective. We hypothesize that in VP3D+{Conf.-Concat}, it is difficult to effectively entangle the position and confidence scores, with inherently different scales and meanings, using a single convolution.

\subsection{Ablation Study}
\begin{table}[t!]
  \caption{Effect of VP3D's~\cite{pavllo2019_videopose3d} receptive field on $\text{MPJPE}_{\leq 0.1}$ and the performance gain compared to lower- or upper-bounds.}
  \label{tab:mpjpe-receptive-field}
  \vspace*{-0.5\baselineskip}
  \centering
    \resizebox{0.85\linewidth}{!}{
    \begin{tabular}{c|l|c}
    \toprule
    \textbf{\begin{tabular}[c]{@{}c@{}}Receptive\\ Field\end{tabular}} & \multicolumn{1}{c|}{\textbf{Model}} & \textbf{\begin{tabular}[c]{@{}c@{}}\textbf{Average}\\ \textbf{MPJPE}$_{\boldsymbol{\leq 0.1}}$\end{tabular}} \\ \hline
    \multirow{4}{*}{1}                                      & VP3D/H36M \textit{(upper bound)}          & 96.87               \\
                                                            & VP3D/H36M+{TAGN}                   & 91.39 \better{($\downarrow 5.48$)} \\ \cline{2-3}
                                                            & VP3D/H36M-C \textit{(lower bound)}& 83.09               \\
                                                            & VP3D/H36M-C + {CA-Conv}    & {80.25} \better{($\downarrow 2.84$)} \\ \hline
    \multirow{4}{*}{9}                                      & VP3D/H36M \textit{(upper bound)}          & 96.26          \\
                                                            & VP3D/H36M+{TAGN}                   & 92.79 \better{($\downarrow 3.47$)} \\ \cline{2-3}
                                                            & VP3D/H36M-C \textit{(upper bound)}                      & 79.37          \\
                                                            & VP3D/H36M-C + {CA-Conv}    & {76.94} \better{($\downarrow 2.43$)}\\ \hline
    \multirow{4}{*}{27}                                     & VP3D/H36M \textit{(upper bound)}          & 94.19  \\
                                                            & VP3D/H36M+{TAGN}                   & 88.54 \better{($\downarrow 5.65$)}         \\ \cline{2-3}
                                                            & VP3D/H36M-C {(lower bound)}& 75.46          \\
                                                            & VP3D/H36M-C + {CA-Conv}    & {73.64} \better{($\downarrow 1.82$)} \\ \hline
    \multirow{4}{*}{81}                                     & VP3D/H36M  \textit{(upper bound)}         & 93.35          \\
                                                            & VP3D/H36M+{TAGN}                   & 87.69 \better{($\downarrow 5.66$)}         \\ \cline{2-3}
                                                            & VP3D/H36M-C \textit{(lower bound)}& 73.61          \\
                                                            & VP3D/H36M-C + {CA-Conv}    & {71.84} \better{($\downarrow 1.77$)} \\ \hline
    \end{tabular}
    }
    \vspace*{-0.5\baselineskip}
\end{table}

In Table \ref{tab:mpjpe-receptive-field}, we evaluate the effect of temporal receptive fields $\{1, 9, 27, 81\}$ on VP3D performance, trained with either TAGN ($\sigma=0.3$, $p=k=50\%$) or CA-Conv ($\gamma=1$), against its baseline. We observe a consistent performance boost for different receptive field sizes when TAGN or CA-Conv are adopted. This demonstrates our approaches are suitable for various choices of this parameter.

\begin{table}[t!]
  \caption{MPJPE$_{\leq 0.1}$ of VP3D \cite{pavllo2019_videopose3d} models trained/tested on 2D keypoints detected by HRNet \cite{sun2019_hrnet} and Lite-HRNet \cite{yu2021_lite-hrnet}.} 
  \label{tab:mpjpe-hrnet-lite-hrnet}
  \vspace*{-0.5\baselineskip}
  \centering
    \resizebox{.95\linewidth}{!}{
    \begin{tabular}{c|l|cc}
    \toprule
\multirow{2}{*}{\textbf{\begin{tabular}[c]{@{}c@{}}Training 2D \\ keypoints\end{tabular}}} & \multicolumn{1}{c|}{\multirow{2}{*}{\textbf{Model}}} & \multicolumn{2}{c}{\textbf{Testing 2D keypoints}}                   \\ \cline{3-4} 
                                                                                      & \multicolumn{1}{c|}{}                                & \multicolumn{1}{c|}{\textbf{HRNet}} & \textbf{Lite-HRNet} \\ \hline
\multirow{4}{*}{\textbf{HRNet}}                                                       & VP3D/H36M \textit{(upper bound)}                    & \multicolumn{1}{c|}{100.50}         & 118.45              \\
                                                                                      & VP3D/H36M+TAGN                                       & \multicolumn{1}{c|}{92.94}          & 112.26              \\
                                                                                      & VP3D/H36M-C \textit{(lower bound)}             & \multicolumn{1}{c|}{77.92}          & 98.03               \\
                                                                                      & VP3D/H36M-C + CA-Conv                        & \multicolumn{1}{c|}{76.13}          & 99.31               \\ \hline
\multirow{4}{*}{\textbf{\begin{tabular}[c]{@{}c@{}}Lite-\\ HRNet \end{tabular}}}      & VP3D/H36M  \textit{(upper bound)}                                           & \multicolumn{1}{c|}{101.66}         & 115.94              \\
                                                                                      & VP3D/H36M+TAGN                                       & \multicolumn{1}{c|}{98.07}          & 110.70              \\
                                                                                      & VP3D/H36M-C \textit{(lower bound)}                      & \multicolumn{1}{c|}{84.08}          & 94.16               \\
                                                                                      & VP3D/H36M-C + CA-Conv                        & \multicolumn{1}{c|}{81.59}          & 89.01               \\ \hline
\end{tabular}
    }
\vspace*{-\baselineskip}
\end{table}

We examine the effect of the 2D detector in Table~\ref{tab:mpjpe-hrnet-lite-hrnet}. In addition to HRNet, we also employ Lite-HRNet \cite{yu2021_lite-hrnet} - a lightweight version of HRNet - which estimates 2D pose in real-time. Due to the simplicity of Lite-HRNet, the detected 2D keypoints are less accurate and generally result in a higher 3D lifting error. Regardless, the models with TAGN and CA-Conv still outperform the upper- and lower-bounds. Moreover, we also do cross-detector evaluations, where the 2D keypoints detector at training differs from the one at test time. When VP3D/H36M-C+CA-Conv is trained with HRNet 2D
pose, it is exposed to more accurate 2D input pose and confidence scores compared to Lite-HRNet (see the Supplementary for a comparison of Lite-HRNet and HRNet confidence score distribution). This slightly affects the performance when tested on Lite-HRNet keypoints but not the other way around. 
In general, CA-Conv expects similar or higher quality test-time 2D keypoints and confidence scores to achieve superior robustness.
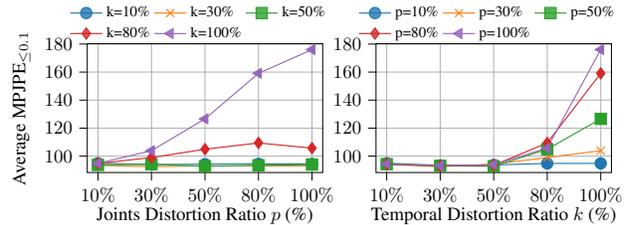
\begin{figure}[t]
    \centering
    \resizebox{\linewidth}{!}{\begin{tikzpicture}

\definecolor{crimson2143940}{RGB}{214,39,40}
\definecolor{darkgray176}{RGB}{176,176,176}
\definecolor{darkorange25512714}{RGB}{255,127,14}
\definecolor{forestgreen4416044}{RGB}{44,160,44}
\definecolor{lightgray204}{RGB}{204,204,204}
\definecolor{mediumpurple148103189}{RGB}{148,103,189}
\definecolor{steelblue31119180}{RGB}{31,119,180}

\begin{groupplot}[group style={group size=2 by 1}]
\nextgroupplot[
height=4cm, width=6cm,
legend style={font=\footnotesize},
legend cell align={left},
legend style={
  fill opacity=0.8,
  draw opacity=1,
  text opacity=1,
  at={(-0.1,1.35)},
  legend columns=3,
  anchor=north west,
  draw=lightgray204,
  draw=none, fill=none
},
tick align=outside,
tick pos=left,
x grid style={darkgray176},
xlabel={Joints Distortion Ratio $p$ (\%)},
xmajorgrids,
xmin=-0.2, xmax=4.2,
xtick style={color=black},
xtick={0,1,2,3,4},
xticklabels={
  \(\displaystyle 10\%\),
  \(\displaystyle 30\%\),
  \(\displaystyle 50\%\),
  \(\displaystyle 80\%\),
  \(\displaystyle 100\%\)
},
y grid style={darkgray176},
ylabel={Average MPJPE\(\displaystyle _{\leq 0.1}\)},
ymajorgrids,
ymin=88.302, ymax=180.218,
ytick style={color=black}
]
\addplot [semithick, steelblue31119180, mark=*, mark size=3, mark options={solid}]
table {%
0 94.96
1 93.96
2 94.28
3 94.48
4 94.46
};
\addlegendentry{k=10\%}
\addplot [semithick, darkorange25512714, mark=x, mark size=3, mark options={solid}]
table {%
0 93.42
1 92.48
2 93.1
3 92.92
4 93.12
};
\addlegendentry{k=30\%}
\addplot [semithick, forestgreen4416044, mark=square*, mark size=3, mark options={solid}]
table {%
0 93.73
1 94.26
2 92.72
3 93.46
4 93.96
};
\addlegendentry{k=50\%}
\addplot [semithick, crimson2143940, mark=diamond*, mark size=3, mark options={solid}]
table {%
0 94.78
1 98.98
2 104.94
3 109.42
4 105.73
};
\addlegendentry{k=80\%}
\addplot [semithick, mediumpurple148103189, mark=triangle*, mark size=3, mark options={solid,rotate=90}]
table {%
0 94.83
1 103.91
2 126.6
3 159.11
4 176.04
};
\addlegendentry{k=100\%}

\nextgroupplot[
legend cell align={left},
legend style={
  fill opacity=0.8,
  draw opacity=1,
  text opacity=1,
  at={(-0.1,1.35)},
  anchor=north west,
  draw=lightgray204,
  draw=none, fill=none,
  legend columns=3
},
height=4cm, width=6cm,
legend style={font=\footnotesize},
tick align=outside,
tick pos=left,
x grid style={darkgray176},
xlabel={Temporal Distortion Ratio $k$ (\%)},
xmajorgrids,
xmin=-0.2, xmax=4.2,
xtick style={color=black},
xtick={0,1,2,3,4},
xticklabels={
  \(\displaystyle 10\%\),
  \(\displaystyle 30\%\),
  \(\displaystyle 50\%\),
  \(\displaystyle 80\%\),
  \(\displaystyle 100\%\)
},
y grid style={darkgray176},
ymajorgrids,
ymin=88.302, ymax=180.218,
ytick style={color=black}
]
\addplot [semithick, steelblue31119180, mark=*, mark size=3, mark options={solid}]
table {%
0 94.96
1 93.42
2 93.73
3 94.78
4 94.83
};
\addlegendentry{p=10\%}
\addplot [semithick, darkorange25512714, mark=x, mark size=3, mark options={solid}]
table {%
0 93.96
1 92.48
2 94.26
3 98.98
4 103.91
};
\addlegendentry{p=30\%}
\addplot [semithick, forestgreen4416044, mark=square*, mark size=3, mark options={solid}]
table {%
0 94.28
1 93.1
2 92.72
3 104.94
4 126.6
};
\addlegendentry{p=50\%}
\addplot [semithick, crimson2143940, mark=diamond*, mark size=3, mark options={solid}]
table {%
0 94.48
1 92.92
2 93.46
3 109.42
4 159.11
};
\addlegendentry{p=80\%}
\addplot [semithick, mediumpurple148103189, mark=triangle*, mark size=3, mark options={solid,rotate=90}]
table {%
0 94.46
1 93.12
2 93.96
3 105.73
4 176.04
};
\addlegendentry{p=100\%}
\end{groupplot}

\end{tikzpicture}}
    \vspace*{-1.5\baselineskip}
    \caption{Effect of TAGN's joint and temporal distortion ratios ($p$ and $k$) with $\sigma=0.3$. Here, the lifter architecture is VP3D.}
    \label{fig:distortion_selection}
    \vspace*{-1.5\baselineskip}
\end{figure}

We analyze the impact of TAGN's temporal and joint distortion ratios (i.e. $k$ and $p$) in Figure \ref{fig:distortion_selection}. While the joint distortion ratio ($p$) does not visibly impact the performance, increasing the temporal distortion ratio ($k$) gradually degrades the results, especially when more than half of the frames are affected. This highlights the effectiveness of TAGN and the fact that blindly adding Gaussian noise to every joint and frame leads to the worst performance since the model cannot rely on any joint or frame within its receptive field.

\vspace*{-0.5\baselineskip}
\section{Conclusion}
\label{sec:conclusion}

By rigorously studying the robustness of 2D-to-3D pose lifters under two introduced benchmarks with video corruptions, we showed in spite of accurate 3D pose estimation of state-of-the-art 2D-to-3D pose lifters on standard benchmarks, they are extremely sensitive to the unexpected distortions in their input. 
This observation prompts the necessity of careful investigation of robustness in this task. 
We took steps towards improving the robustness of lifter models by introducing (1) TAGN - a simple 2D pose augmentation approach, (2) infusing the knowledge of the 2D detector's confidence score at the 3D lifter side through our confidence-aware convolution block design. Our extensive experimental results revealed the effectiveness of our methods in boosting the robustness of 3D pose lifters.

\section*{Acknowledgement}
This project has been funded by the Jump ARCHES endowment through the Health Care Engineering Systems Center, JSPS/MEXT KAKENHI (24K20830), and ROIS NII Open Collaborative Research (2023-23FC01, 2024-24S1201).

{\small
\bibliographystyle{ieee_fullname}
\bibliography{egbib}

\begin{thebibliography}{10}\itemsep=-1pt

\bibitem{aliakbarian2022_flag}
Sadegh Aliakbarian, Pashmina Cameron, Federica Bogo, Andrew Fitzgibbon, and Tom Cashman.
\newblock Flag: Flow-based 3{D} avatar generation from sparse observations.
\newblock In {\em 2022 Computer Vision and Pattern Recognition}, June 2022.

\bibitem{9506392}
Amin Ansarian and Maria~A. Amer.
\newblock Realistic augmentation for effective 2{D} human pose estimation under occlusion.
\newblock In {\em 2021 IEEE International Conference on Image Processing (ICIP)}, pages 919--923, 2021.

\bibitem{biggs20203d}
Benjamin Biggs, David Novotny, Sebastien Ehrhardt, Hanbyul Joo, Ben Graham, and Andrea Vedaldi.
\newblock 3{D} multi-bodies: Fitting sets of plausible 3{D} human models to ambiguous image data.
\newblock {\em Advances in Neural Information Processing Systems}, 33:20496--20507, 2020.

\bibitem{Chen_2017_CVPR}
Ching-Hang Chen and Deva Ramanan.
\newblock 3{D} human pose estimation = 2{D} pose estimation + matching.
\newblock In {\em Proceedings of the IEEE Conference on Computer Vision and Pattern Recognition (CVPR)}, July 2017.

\bibitem{chen2018_cpn}
Yilun Chen, Zhicheng Wang, Yuxiang Peng, Zhiqiang Zhang, Gang Yu, and Jian Sun.
\newblock Cascaded pyramid network for multi-person pose estimation.
\newblock In {\em Proceedings of the IEEE Conference on Computer Vision and Pattern Recognition (CVPR)}, June 2018.

\bibitem{cheng20203d}
Yu Cheng, Bo Yang, Bo Wang, and Robby~T Tan.
\newblock 3{D} human pose estimation using spatio-temporal networks with explicit occlusion training.
\newblock In {\em Proceedings of the AAAI Conference on Artificial Intelligence}, volume~34, pages 10631--10638, 2020.

\bibitem{9010921}
Yu Cheng, Bo Yang, Bo Wang, Yan Wending, and Robby Tan.
\newblock Occlusion-aware networks for 3{D} human pose estimation in video.
\newblock In {\em 2019 IEEE/CVF International Conference on Computer Vision (ICCV)}, pages 723--732, 2019.

\bibitem{czarnecki2013_uncertaintymeasure}
Wojciech~M. Czarnecki and Igor~T. Podolak.
\newblock Machine learning with known input data uncertainty measure.
\newblock In Khalid Saeed, Rituparna Chaki, Agostino Cortesi, and S{\l}awomir Wierzcho{\'{n}}, editors, {\em Computer Information Systems and Industrial Management}, pages 379--388, Berlin, Heidelberg, 2013. Springer Berlin Heidelberg.

\bibitem{gong2021poseaug}
Kehong Gong, Jianfeng Zhang, and Jiashi Feng.
\newblock Poseaug: A differentiable pose augmentation framework for 3{D} human pose estimation.
\newblock In {\em Proceedings of the IEEE/CVF Conference on Computer Vision and Pattern Recognition}, pages 8575--8584, 2021.

\bibitem{hendrycks2018benchmarking}
Dan Hendrycks and Thomas Dietterich.
\newblock Benchmarking neural network robustness to common corruptions and perturbations.
\newblock In {\em International Conference on Learning Representations}, 2019.

\bibitem{hoangzehni2022_dne}
Trung-Hieu Hoang, Mona Zehni, Huaijin Xu, George Heintz, Christopher Zallek, and Minh~N. Do.
\newblock Towards a comprehensive solution for a vision-based digitized neurological examination.
\newblock {\em IEEE Journal of Biomedical and Health Informatics}, 26(8):4020--4031, 2022.

\bibitem{hu2021conditional}
Wenbo Hu, Changgong Zhang, Fangneng Zhan, Lei Zhang, and Tien-Tsin Wong.
\newblock Conditional directed graph convolution for 3{D} human pose estimation.
\newblock In {\em Proceedings of the 29th ACM International Conference on Multimedia}, pages 602--611, 2021.

\bibitem{h36m_pami}
Catalin Ionescu, Dragos Papava, Vlad Olaru, and Cristian Sminchisescu.
\newblock Human3.6{M}: Large scale datasets and predictive methods for 3{D} human sensing in natural environments.
\newblock {\em IEEE Transactions on Pattern Analysis and Machine Intelligence}, 36(7):1325--1339, jul 2014.

\bibitem{jiang2021skeletor}
Tao Jiang, Necati~Cihan Camgoz, and Richard Bowden.
\newblock Skeletor: Skeletal transformers for robust body-pose estimation.
\newblock In {\em Proceedings of the IEEE/CVF Conference on Computer Vision and Pattern Recognition}, pages 3394--3402, 2021.

\bibitem{kamann2021_semantic}
Christoph Kamann and Carsten Rother.
\newblock Benchmarking the robustness of semantic segmentation models with respect to common corruptions.
\newblock {\em Int. J. Comput. Vis.}, 129(2):462--483, 2021.

\bibitem{kocabas2021pare}
Muhammed Kocabas, Chun-Hao~P Huang, Otmar Hilliges, and Michael~J Black.
\newblock {PARE}: Part attention regressor for 3{D} human body estimation.
\newblock In {\em Proceedings of the IEEE/CVF International Conference on Computer Vision}, pages 11127--11137, 2021.

\bibitem{li2014}
Sijin Li and Antoni~B Chan.
\newblock 3{D} human pose estimation from monocular images with deep convolutional neural network.
\newblock In {\em Asian Conference on Computer Vision}, pages 332--347. Springer, 2014.

\bibitem{Li_2020_CVPR}
Shichao Li, Lei Ke, Kevin Pratama, Yu-Wing Tai, Chi-Keung Tang, and Kwang-Ting Cheng.
\newblock Cascaded deep monocular 3{D} human pose estimation with evolutionary training data.
\newblock In {\em Proceedings of the IEEE/CVF Conference on Computer Vision and Pattern Recognition (CVPR)}, June 2020.

\bibitem{li2022exploiting}
Wenhao Li, Hong Liu, Runwei Ding, Mengyuan Liu, Pichao Wang, and Wenming Yang.
\newblock Exploiting temporal contexts with strided transformer for 3{D} human pose estimation.
\newblock {\em IEEE Transactions on Multimedia}, 2022.

\bibitem{lin2014_coco}
Tsung-Yi Lin, Michael Maire, Serge Belongie, James Hays, Pietro Perona, Deva Ramanan, Piotr Doll{\'a}r, and C.~Lawrence Zitnick.
\newblock Microsoft {COCO}: {C}ommon {O}bjects in {C}ontext.
\newblock In David Fleet, Tomas Pajdla, Bernt Schiele, and Tinne Tuytelaars, editors, {\em Computer Vision -- ECCV 2014}, pages 740--755, Cham, 2014. Springer International Publishing.

\bibitem{liu2020_attention}
Ruixu Liu, Ju Shen, He Wang, Chen Chen, Sen-ching Cheung, and Vijayan Asari.
\newblock Attention mechanism exploits temporal contexts: Real-time 3{D} human pose reconstruction.
\newblock In {\em Proceedings of the IEEE/CVF Conference on Computer Vision and Pattern Recognition}, pages 5064--5073, 2020.

\bibitem{luvizon2018}
Diogo~C. Luvizon, David Picard, and Hedi Tabia.
\newblock 2{D}/3{D} pose estimation and action recognition using multitask deep learning.
\newblock In {\em 2018 IEEE/CVF Conference on Computer Vision and Pattern Recognition}, pages 5137--5146, 2018.

\bibitem{martinez2017simple}
Julieta Martinez, Rayat Hossain, Javier Romero, and James~J Little.
\newblock A simple yet effective baseline for 3{D} human pose estimation.
\newblock In {\em Proceedings of the IEEE international conference on computer vision}, pages 2640--2649, 2017.

\bibitem{mehta2017-3dhp}
Dushyant Mehta, Helge Rhodin, Dan Casas, Pascal Fua, Oleksandr Sotnychenko, Weipeng Xu, and Christian Theobalt.
\newblock Monocular 3{D} human pose estimation in the wild using improved cnn supervision.
\newblock In {\em 3{D} Vision (3{D}V), 2017 Fifth International Conference on}. IEEE, 2017.

\bibitem{mono-3dhp2017}
Dushyant Mehta, Helge Rhodin, Dan Casas, Pascal Fua, Oleksandr Sotnychenko, Weipeng Xu, and Christian Theobalt.
\newblock Monocular 3{D} human pose estimation in the wild using improved cnn supervision.
\newblock In {\em 3{D} Vision (3{D}V), 2017 Fifth International Conference on}. IEEE, 2017.

\bibitem{mehta2017_vnect}
Dushyant Mehta, Srinath Sridhar, Oleksandr Sotnychenko, Helge Rhodin, Mohammad Shafiei, Hans-Peter Seidel, Weipeng Xu, Dan Casas, and Christian Theobalt.
\newblock Vnect: Real-time 3{D} human pose estimation with a single {RGB} camera.
\newblock In {\em ACM Transactions on Graphics}, volume~36, July 2017.

\bibitem{michaelis2019dragon}
Claudio Michaelis, Benjamin Mitzkus, Robert Geirhos, Evgenia Rusak, Oliver Bringmann, Alexander~S. Ecker, Matthias Bethge, and Wieland Brendel.
\newblock Benchmarking robustness in object detection: Autonomous driving when winter is coming.
\newblock {\em arXiv preprint arXiv:1907.07484}, 2019.

\bibitem{moreno2017}
Francesc Moreno-Noguer.
\newblock 3{D} human pose estimation from a single image via distance matrix regression.
\newblock In {\em Proceedings of the IEEE conference on computer vision and pattern recognition}, pages 2823--2832, 2017.

\bibitem{park2016}
Sungheon Park, Jihye Hwang, and Nojun Kwak.
\newblock 3{D} human pose estimation using convolutional neural networks with 2{D} pose information.
\newblock In {\em European Conference on Computer Vision}, pages 156--169. Springer, 2016.

\bibitem{park20163d}
Sungheon Park, Jihye Hwang, and Nojun Kwak.
\newblock 3{D} human pose estimation using convolutional neural networks with 2{D} pose information.
\newblock In {\em European Conference on Computer Vision}, pages 156--169. Springer, 2016.

\bibitem{pavlakos2017coarse}
Georgios Pavlakos, Xiaowei Zhou, Konstantinos~G Derpanis, and Kostas Daniilidis.
\newblock Coarse-to-fine volumetric prediction for single-image 3{D} human pose.
\newblock In {\em Proceedings of the IEEE conference on computer vision and pattern recognition}, pages 7025--7034, 2017.

\bibitem{pavllo2019_videopose3d}
Dario Pavllo, Christoph Feichtenhofer, David Grangier, and Michael Auli.
\newblock 3{D} human pose estimation in video with temporal convolutions and semi-supervised training.
\newblock In {\em Proceedings of the IEEE/CVF Conference on Computer Vision and Pattern Recognition}, pages 7753--7762, 2019.

\bibitem{10.5555/1462129}
Joaquin Quionero-Candela, Masashi Sugiyama, Anton Schwaighofer, and Neil~D. Lawrence.
\newblock {\em Dataset Shift in Machine Learning}.
\newblock The MIT Press, 2009.

\bibitem{7301338}
Umer Rafi, Juergen Gall, and Bastian Leibe.
\newblock A semantic occlusion model for human pose estimation from a single depth image.
\newblock In {\em 2015 IEEE Conference on Computer Vision and Pattern Recognition Workshops (CVPRW)}, pages 67--74, 2015.

\bibitem{reed1992_jittered}
Russell Reed, Se young Oh, and Robert~J. Marks.
\newblock Regularization using jittered training data.
\newblock In {\em Proceedings of IJCNN International Joint Conference on Neural Networks}, volume~3, pages 147--152 vol.3, 1992.

\bibitem{rempe2021humor}
Davis Rempe, Tolga Birdal, Aaron Hertzmann, Jimei Yang, Srinath Sridhar, and Leonidas~J. Guibas.
\newblock Hu{M}o{R}: 3{D} human motion model for robust pose estimation.
\newblock In {\em International Conference on Computer Vision (ICCV)}, 2021.

\bibitem{rockwell2020full}
Chris Rockwell and David~F Fouhey.
\newblock Full-body awareness from partial observations.
\newblock In {\em European Conference on Computer Vision}, pages 522--539. Springer, 2020.

\bibitem{sarandiIROSWS18}
Istv\'{a}n S\'{a}r\'{a}ndi, Timm Linder, Kai~Oliver Arras, and Bastian Leibe.
\newblock How robust is 3{D} human pose estimation to occlusion?
\newblock IEEE/RSJ International Conference on Intelligent Robots and Systems (IROS'18) - Workshop on Robotic Co-workers 4.0: Human Safety and Comfort in Human-Robot Interactive Social Environments, October 2018.

\bibitem{sarandi2018synthetic}
Istv{\'a}n S{\'a}r{\'a}ndi, Timm Linder, Kai~O Arras, and Bastian Leibe.
\newblock Synthetic occlusion augmentation with volumetric heatmaps for the 2018 {ECCV} posetrack challenge on 3{D} human pose estimation.
\newblock {\em arXiv preprint arXiv:1809.04987}, 2018.

\bibitem{shan2021improving}
Wenkang Shan, Haopeng Lu, Shanshe Wang, Xinfeng Zhang, and Wen Gao.
\newblock Improving robustness and accuracy via relative information encoding in 3{D} human pose estimation.
\newblock In {\em Proceedings of the 29th ACM International Conference on Multimedia}, pages 3446--3454, 2021.

\bibitem{sharma2019monocular}
Saurabh Sharma, Pavan~Teja Varigonda, Prashast Bindal, Abhishek Sharma, and Arjun Jain.
\newblock Monocular 3{D} human pose estimation by generation and ordinal ranking.
\newblock In {\em Proceedings of the IEEE/CVF international conference on computer vision}, pages 2325--2334, 2019.

\bibitem{sigal2010_humaneva}
Leonid Sigal, Alexandru~O. Balan, and Michael~J. Black.
\newblock {HumanEva}: Synchronized video and motion capture dataset and baseline algorithm for evaluation of articulated human motion.
\newblock {\em International Journal of Computer Vision}, 87(1):4--27, Mar. 2010.

\bibitem{s21217315}
Jan Stenum, Kendra~M Cherry-Allen, Connor~O Pyles, Rachel~D Reetzke, Michael~F Vignos, and Ryan~T Roemmich.
\newblock {Applications of Pose Estimation in Human Health and Performance across the Lifespan}.
\newblock {\em Sensors}, 21(21), 2021.

\bibitem{sun2019_hrnet}
Ke Sun, Bin Xiao, Dong Liu, and Jingdong Wang.
\newblock Deep high-resolution representation learning for human pose estimation.
\newblock In {\em CVPR}, 2019.

\bibitem{sun2018integral}
Xiao Sun, Bin Xiao, Fangyin Wei, Shuang Liang, and Yichen Wei.
\newblock Integral human pose regression.
\newblock In {\em Proceedings of the European conference on computer vision (ECCV)}, pages 529--545, 2018.

\bibitem{sun2021monocular}
Yu Sun, Qian Bao, Wu Liu, Yili Fu, Michael~J Black, and Tao Mei.
\newblock Monocular, one-stage, regression of multiple 3{D} people.
\newblock In {\em Proceedings of the IEEE/CVF International Conference on Computer Vision}, pages 11179--11188, 2021.

\bibitem{sun2022putting}
Yu Sun, Wu Liu, Qian Bao, Yili Fu, Tao Mei, and Michael~J Black.
\newblock Putting people in their place: Monocular regression of 3{D} people in depth.
\newblock In {\em Proceedings of the IEEE/CVF Conference on Computer Vision and Pattern Recognition}, pages 13243--13252, 2022.

\bibitem{Tekin2017LearningTF}
Bugra Tekin, Pablo M{\'a}rquez-Neila, Mathieu Salzmann, and P. Fua.
\newblock Learning to fuse 2{D} and 3{D} image cues for monocular body pose estimation.
\newblock {\em 2017 IEEE International Conference on Computer Vision (ICCV)}, pages 3961--3970, 2017.

\bibitem{tekin2016direct}
Bugra Tekin, Artem Rozantsev, Vincent Lepetit, and Pascal Fua.
\newblock Direct prediction of 3{D} body poses from motion compensated sequences.
\newblock In {\em Proceedings of the IEEE Conference on Computer Vision and Pattern Recognition}, pages 991--1000, 2016.

\bibitem{Tome_2017_CVPR}
Denis Tome, Chris Russell, and Lourdes Agapito.
\newblock Lifting from the deep: Convolutional 3{D} pose estimation from a single image.
\newblock In {\em Proceedings of the IEEE Conference on Computer Vision and Pattern Recognition (CVPR)}, July 2017.

\bibitem{wandt2022elepose}
Bastian Wandt, James~J Little, and Helge Rhodin.
\newblock Elepose: Unsupervised 3{D} human pose estimation by predicting camera elevation and learning normalizing flows on 2{D} poses.
\newblock In {\em Proceedings of the IEEE/CVF Conference on Computer Vision and Pattern Recognition}, pages 6635--6645, 2022.

\bibitem{wang2021_show}
Danding Wang, Wencan Zhang, and Brian Lim.
\newblock Show or suppress? managing input uncertainty in machine learning model explanations.
\newblock {\em Artificial Intelligence}, 294:103456, 01 2021.

\bibitem{wang2021_robustness}
Jiahang Wang, Sheng Jin, Wentao Liu, Weizhong Liu, Chen Qian, and Ping Luo.
\newblock When human pose estimation meets robustness: Adversarial algorithms and benchmarks.
\newblock In {\em 2021 IEEE/CVF Conference on Computer Vision and Pattern Recognition (CVPR)}, pages 11850--11859, Los Alamitos, CA, USA, jun 2021. IEEE Computer Society.

\bibitem{WANG2021}
Jinbao Wang, Shujie Tan, Xiantong Zhen, Shuo Xu, Feng Zheng, Zhenyu He, and Ling Shao.
\newblock Deep 3{D} human pose estimation: A review.
\newblock {\em Computer Vision and Image Understanding}, 210:103225, 2021.

\bibitem{wang2020motion}
Jingbo Wang, Sijie Yan, Yuanjun Xiong, and Dahua Lin.
\newblock Motion guided 3{D} pose estimation from videos.
\newblock In {\em European Conference on Computer Vision}, pages 764--780. Springer, 2020.

\bibitem{wei2016cpm}
Shih-En Wei, Varun Ramakrishna, Takeo Kanade, and Yaser Sheikh.
\newblock Convolutional pose machines.
\newblock In {\em CVPR}, 2016.

\bibitem{wu2019_detectron2}
Yuxin Wu, Alexander Kirillov, Francisco Massa, Wan-Yen Lo, and Ross Girshick.
\newblock Detectron2.
\newblock \url{https://github.com/facebookresearch/detectron2}, 2019.

\bibitem{yan2018_stgc}
Sijie Yan, Yuanjun Xiong, and Dahua Lin.
\newblock Spatial temporal graph convolutional networks for skeleton-based action recognition.
\newblock In {\em Proceedings of the Thirty-Second AAAI Conference on Artificial Intelligence and Thirtieth Innovative Applications of Artificial Intelligence Conference and Eighth AAAI Symposium on Educational Advances in Artificial Intelligence}. AAAI Press, 2018.

\bibitem{yu2021_lite-hrnet}
Changqian Yu, Bin Xiao, Changxin Gao, Lu Yuan, Lei Zhang, Nong Sang, and Jingdong Wang.
\newblock Lite-{HR}net: A lightweight high-resolution network.
\newblock In {\em CVPR}, 2021.

\bibitem{zeng2020_srnet}
Ailing Zeng, Xiao Sun, Fuyang Huang, Minhao Liu, Qiang Xu, and Stephen Ching-Feng Lin.
\newblock {SRN}et: Improving generalization in 3{D} human pose estimation with a split-and-recombine approach.
\newblock In {\em ECCV}, 2020.

\bibitem{zeng2021learning}
Ailing Zeng, Xiao Sun, Lei Yang, Nanxuan Zhao, Minhao Liu, and Qiang Xu.
\newblock Learning skeletal graph neural networks for hard 3{D} pose estimation.
\newblock In {\em Proceedings of the IEEE/CVF International Conference on Computer Vision}, pages 11436--11445, 2021.

\bibitem{9157481}
Tianshu Zhang, Buzhen Huang, and Yangang Wang.
\newblock Object-occluded human shape and pose estimation from a single color image.
\newblock In {\em 2020 IEEE/CVF Conference on Computer Vision and Pattern Recognition (CVPR)}, pages 7374--7383, 2020.

\bibitem{openpose}
Tomas~Simon Zhe~Cao, Gines~Hidalgo and Yaser~Sheikh Shih-En~Wei.
\newblock Openpose: Realtime multi-person 2{D} pose estimation using part affinity fields.
\newblock {\em IEEE Transactions on Pattern Analysis and Machine Intelligence}, 2019.

\bibitem{zheng2021_poseformer}
Ce Zheng, Sijie Zhu, Matias Mendieta, Taojiannan Yang, Chen Chen, and Zhengming Ding.
\newblock 3{D} human pose estimation with spatial and temporal transformers.
\newblock {\em Proceedings of the IEEE International Conference on Computer Vision (ICCV)}, 2021.

\bibitem{zheng20213d}
Ce Zheng, Sijie Zhu, Matias Mendieta, Taojiannan Yang, Chen Chen, and Zhengming Ding.
\newblock 3{D} human pose estimation with spatial and temporal transformers.
\newblock In {\em Proceedings of the IEEE/CVF International Conference on Computer Vision}, pages 11656--11665, 2021.

\bibitem{zhou2017towards}
Xingyi Zhou, Qixing Huang, Xiao Sun, Xiangyang Xue, and Yichen Wei.
\newblock Towards 3{D} human pose estimation in the wild: a weakly-supervised approach.
\newblock In {\em Proceedings of the IEEE International Conference on Computer Vision}, pages 398--407, 2017.

\bibitem{zhou2016sparseness}
Xiaowei Zhou, Menglong Zhu, Spyridon Leonardos, Konstantinos~G Derpanis, and Kostas Daniilidis.
\newblock Sparseness meets deepness: 3{D} human pose estimation from monocular video.
\newblock In {\em Proceedings of the IEEE conference on computer vision and pattern recognition}, pages 4966--4975, 2016.

\end{thebibliography}
}

\end{document}